\documentclass[11pt]{article}

\usepackage[preprint]{acl}

\usepackage{times}
\usepackage{latexsym}

\usepackage[T1]{fontenc}
\usepackage[utf8]{inputenc}

\usepackage{microtype}

\usepackage{inconsolata}

\usepackage{graphicx}

\usepackage{makecell}

\usepackage{enumitem}
\usepackage{lipsum}
\usepackage{graphicx}
\usepackage{booktabs}
\usepackage{multirow}
\usepackage{comment}
\usepackage{afterpage}
\usepackage{multirow}
\usepackage{floatrow}
\newfloatcommand{capbtabbox}{table}[][\FBwidth]

\usepackage{amsmath}
\usepackage{courier}
\usepackage[T1]{fontenc}
\usepackage{underscore}

\usepackage[dvipsnames]{xcolor}   
\usepackage{booktabs}
\usepackage{array}        % 可选，增强表格功能
\usepackage{multirow}
\usepackage{makecell}
\usepackage{caption}

\usepackage[utf8]{inputenc}
\usepackage[T1]{fontenc}
\usepackage[most]{tcolorbox}
\usepackage{amssymb}

\usepackage[table,dvipsnames]{xcolor}

\newcolumntype{C}[1]{>{\centering\arraybackslash}p{#1}}
\newcolumntype{L}[1]{>{\raggedright\arraybackslash}p{#1}}
\newcolumntype{R}[1]{>{\raggedleft\arraybackslash}p{#1}}

\definecolor{lightpink}{RGB}{255,228,225} 
\definecolor{lightgreen}{RGB}{224,255,224}

\newcommand{\duo}[1]{{\color{black}#1}}

\newcommand{\zm}[1]{{\color{black}#1}}
\newcommand{\re}[1]{{\color{black}#1}}

\usepackage{tabularx}

\usepackage{amsmath, amssymb}

\usepackage[most]{tcolorbox}
\usepackage{soul}
\usepackage{xcolor}
\definecolor{hlgray}{gray}{0.9}
\sethlcolor{hlgray} % 设置 soul 的高亮颜色为浅灰

\newtcbox{\hlemerald}[1][]{%
    on line, 
    arc=0pt,                          % 微圆角，显得更专业
    colback=Emerald!15,               % 比背景深一点点，突出显示
    colframe=white,           % 与外层 box 的边框色一致
    boxrule=0pt,                    % 极细边框
    boxsep=0pt, 
    left=2pt, right=2pt, top=2pt, bottom=2pt, % 紧凑内边距
    #1
}

\newtcbox{\hlgray}[1][]{%
    on line, 
    arc=0pt,                          % 微圆角，显得更专业
    colback=gray!15,               % 比背景深一点点，突出显示
    colframe=white,           % 与外层 box 的边框色一致
    boxrule=0pt,                    % 极细边框
    boxsep=0pt, 
    outer arc=0pt,
    width=\linewidth,
    breakable,
    left=2pt, right=2pt, top=2pt, bottom=2pt, % 紧凑内边距
    #1
}
\title{Collaborative Belief Reasoning with LLMs for Efficient \\ Multi-Agent Collaboration}

\author{
 \textbf{Zhimin Wang\textsuperscript{$\heartsuit$}}, 
 \textbf{Duo Wu\textsuperscript{$\heartsuit$}},
 \textbf{Shaokang He\textsuperscript{$\heartsuit$}}
 ,
 \textbf{Linjia Kang\textsuperscript{$\heartsuit$}},
 \textbf{Jinghe Wang\textsuperscript{$\heartsuit$}}, \\
 \textbf{Jing Yu\textsuperscript{$\clubsuit$}},
 \textbf{Kai Zhu\textsuperscript{$\spadesuit$}},
 \textbf{Jiawei Li\textsuperscript{$\spadesuit$}},
 \textbf{Zhi Wang \textsuperscript{$\heartsuit$}}, \\
 \textsuperscript{$\heartsuit$}Shenzhen International Graduate School, Tsinghua University, \\
 \textsuperscript{$\spadesuit$}Huawei Tenchnologies Co., Ltd.,
 \textsuperscript{$\clubsuit$}University of Science and Technology of China
}

\begin{document}
\maketitle

%%%%%%%%%%%%%%%%%%%%%%%%%%%%%%%%%%%%%%%

\begin{abstract}
Effective real-world multi-agent collaboration requires not only accurate planning but also the ability to reason about collaborators' intents--a crucial capability for avoiding miscoordination and redundant communication under partial observable environments. Due to their strong planning and reasoning capabilities, large language models (LLMs)  have emerged as promising autonomous agents for collaborative task solving. 
However, existing collaboration frameworks for LLMs overlook their reasoning potential for \textit{dynamic intent inference}, and thus produce inconsistent plans and redundant communication, reducing collaboration efficiency. 
To bridge this gap, we propose \textit{\textbf{CoBel-World}}, a novel framework that equips LLM agents with a \textit{\textbf{Co}llaborative \textbf{Bel}ief \textbf{World}}--an internal representation jointly modeling the physical environment and collaborators' mental states. 
CoBel-World enables agents to parse external open-world knowledge into structured beliefs via a symbolic belief representation module, and perform zero-shot Bayesian-style belief updates through LLM reasoning. This allows agents to proactively detect potential miscoordination (e.g., conflicting plans) and communicate adaptively. Evaluated on challenging embodied benchmarks (i.e., TDW-MAT and C-WAH),  CoBel-World significantly reduces communication cost by \textbf{64-79\%} and improves task completion efficiency by \textbf{4-28\%} compared to the strongest baseline. Our results show that explicit, intent-aware belief modeling is essential for efficient and human-like collaboration in LLM-based multi-agent systems.
\end{abstract}

%%%%%%%%%%%%%%%%%%%%%%%%%%%%%%%%%%%%%%%

%%%%%%%%%%%%%%%%%%%%%%%%%%%%%%%%%%%%%%%
\section{Introduction}
\label{introduction}
% Collaboration is a fundamental social mechanism through which humans solve complex tasks and reshape their environments. 
% In recent years, large language models (LLMs) have demonstrated remarkable capabilities in reasoning, planning, and decision-making \citep{liu2024deepseek,openai2023gpt,comanici2025gemini,wu2025catp}, suggesting growing potential for LLMs to act as autonomous agents capable of participating in collaborative problem-solving. While these advances are promising, the effectiveness of existing LLM-based collaboration frameworks has been largely confined to simple text-based domains with high environmental certainty \citep{hong2023metagpt}. In contrast, real-world collaboration requires agents to coordinate actions under uncertainty and adapt to dynamic, partially observable environments. This raises a key question: Can LLMs, when grounded in the physical world, autonomously coordinate with other agents for effective and efficient collaboration?

% We investigate this question in the context of decentralized embodied multi-agent tasks \citep{Zhang2023BuildingCE,Nayak2024LongHorizonPF,Kannan2023SMARTLLMSM}, where agents must perceive, plan, and act under partial observation \citep{Spaan2006DecentralizedPU,Bernstein2000TheCO}, long-horizon dependencies, and environmental dynamics. In such settings, the primary challenge stems from incomplete and misaligned information across agents \citep{Bernstein2000TheCO,foerster2019bayesian}. Communication thus becomes essential for synchronizing internal states, sharing observations, and aligning intents.

\duo{In recent years, large language models (LLMs) have demonstrated remarkable capabilities in reasoning, planning, and decision-making \citep{liu2024deepseek,openai2023gpt,comanici2025gemini,wu2025large,wu2025catp}, highlighting their growing potential to act as autonomous agents  in collaborative problem-solving. While these advances are promising, the effectiveness of existing LLM-based collaboration frameworks has been largely confined to simple text-based domains with high environmental certainty~\citep{hong2023metagpt,qian2024chatdev,li2023camel}. Real-world collaboration, by contrast, requires agents to coordinate actions under uncertainty and adapt to dynamic, partially observable environments characterized by incomplete and misaligned information~\citep{Bernstein2000TheCO,foerster2019bayesian}. In such scenarios, communication becomes essential for synchronizing internal states, sharing observations, and aligning intents across agents~\citep{pan2025multiagent,chan2023chateval,han2024llm,chen2025optima}.}

\begin{figure*}[t]
    \centering
    \includegraphics[width=0.98\linewidth]{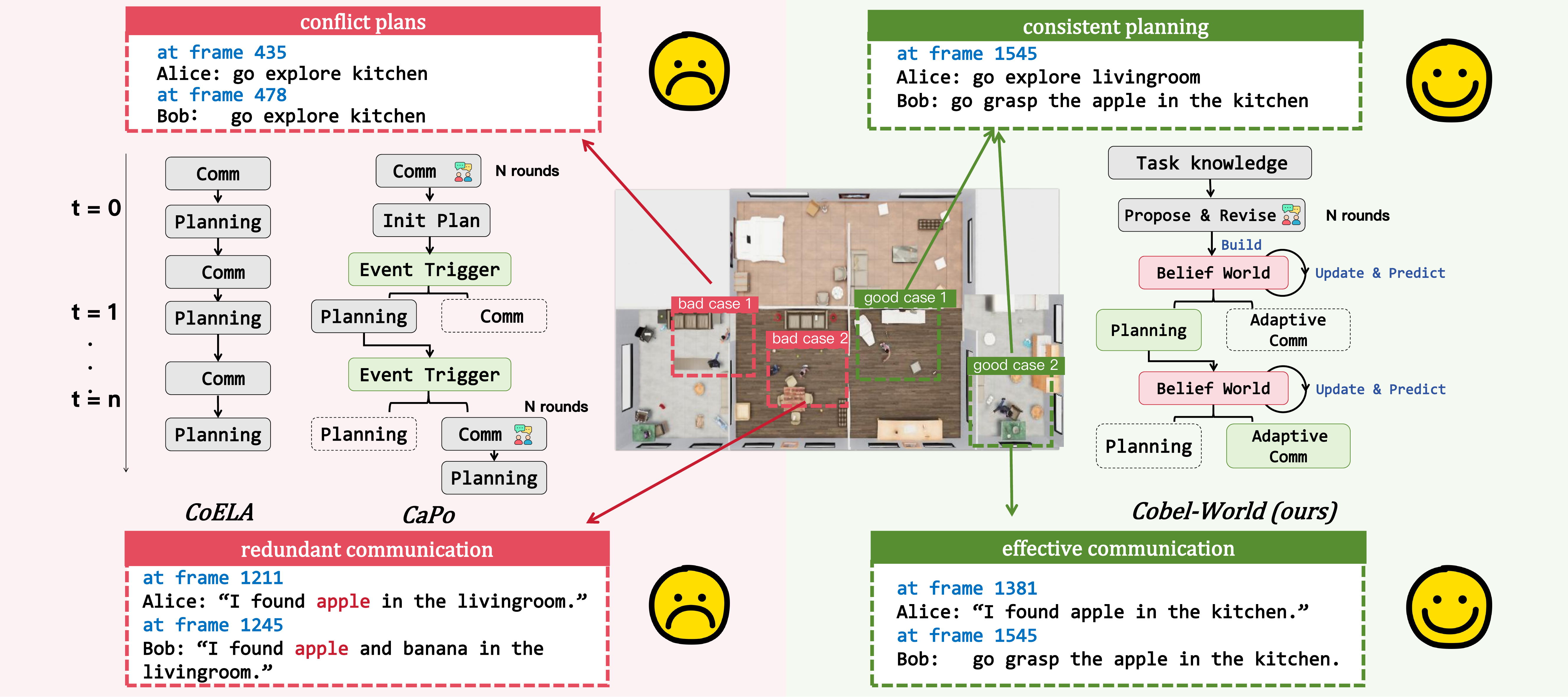} % 替换为你的图片路径
    \vspace{-0.2cm}
    \caption{
    % \textbf{Comparison of existing works with our work.}
    \duo{\textbf{Comparison of existing communication protocols for multi-agent collaboration with our work.}}
    From left to right: 
    % (a) CoELA~\citep{Zhang2023BuildingCE}: A collaboration framework based on step-by-step templated message generation and planning. 
    \duo{(a) CoELA~\citep{Zhang2023BuildingCE}: Fixed templates for step-by-step message generation and planning. }
    % (b) CaPo~\citep{Liu2024CaPoCP}: A collaboration framework based on event-driven multi-round discussion. 
    \duo{(b) CaPo~\citep{Liu2024CaPoCP}: Event-driven multi-round discussion. }
    % (c) Our CoBel-World framework, featuring belief modeling and adaptive collaboration. 
    \duo{(c) CoBel-World (ours): Belief modeling and adaptive collaboration. }
    Our method enables consistent planning and effective communication.}
    \label{fig:first_figure}
% \vspace{-0.3cm}
\end{figure*}

As shown in Figure \ref{fig:first_figure}, recent approaches have explored various communication protocols to enable information sharing and consensus in multi-agent systems. However, these methods typically rely on predefined collaboration schemes and fixed communication protocols-such as step-by-step message generation~\citep{Zhang2023BuildingCE}, dense discussion~\citep{mandi2024roco}, or event-triggered multi-round discussion~\citep{Liu2024CaPoCP}. Crucially, they lack the ability to dynamically identify potential miscoordination and communicate adatively. As a result, \duo{repetitive}  communication and inconsistent planning \duo{frequently occur}, leading to high communication costs and redundant physical actions. 
% These limitations hinder scalability in large-scale, communication-constrained, or human-AI collaborative environments.

% We argue that this shortcoming arises from the absence of belief modeling. In multi-agent systems, beliefs refer to an agent’s internal representation of possible states--including the external environment and the mental states (e.g., intents, knowledge) of collaborators~\citep{kominis2015beliefs,geffner2013concise}. 
\duo{We argue that this limitation stems from the lack of explicit belief modeling. In multi-agent systems, beliefs refer to the agent’s internal representation of the world, including the external  environment and  mental states (e.g., intents, knowledge) of collaborators~\citep{kominis2015beliefs,geffner2013concise}. }
In decentralized multi-agent reinforcement learning (DEC-MARL), belief modeling has proven critical for collaboration under partial observation, enabling agents to infer and align with others’ internal states~\citep{pritz2025belief,wen2019probabilistic,zhai2023dynamic}. With accurate belief estimation, agents can selectively communicate only the valuable information to achieve efficient communication and reach consensus, thus promoting consistent collaboration.

% Despite its advantages, modeling belief for LLM-driven agents presents two primary challenges:

% \begin{itemize}

% \item \textbf{Formulating beliefs in open-ended environments.} 
% Traditional MARL agents operate in low-dimensional, structured environments (e.g., grid worlds) with discrete action space, enabling straightforward belief representation. In contrast, LLM-based embodied agents interact with open-ended physical environments characterized by high-dimensional,  compositional actions, and free-form communication. These features complicate the grounding of linguistic instructions into structured, explicit belief representation.

% \item \textbf{Zero-shot construction of belief models.} In abstract domains like grid-world games \citep{moreno2021neural}, agents are trained on large-scale interaction datasets to infer others’ intents. However, collecting real-world interaction trajectories for fine-tuning LLM agents is prohibitively expensive and often impractical. Moreover, data-driven models may struggle to generalize across diverse, unseen scenarios. This necessitates a zero-shot approach: LLM agents must construct and update beliefs without access to annotated interaction data during pretraining or downstream adaptation.
% %an internal representation of external world under multi-agent interaction.
% \end{itemize}

\duo{Despite its advantages, belief modeling for LLM agents faces two fundamental challenges. First, LLM agents operate in open-ended physical environments featured with high-dimensional, compositional action space and free-form natural language communication. These characteristics make it difficult to ground linguistic instructions into structured, explicit belief representations. Second, collecting real-world interaction trajectories to fine-tune LLMs for inferring others’ intents is prohibitively expensive and often infeasible. Consequently, LLM agents must construct and update beliefs in a zero-shot manner, without access to annotated interaction data during pretraining or downstream adaptation.}

To address these challenges, we propose \textit{\textbf{CoBel-World}}, a novel framework that equips LLM agents with a \textit{\textbf{Co}llaborative \textbf{Bel}ief \textbf{World}}--an internal representation
of the external \duo{environment} and mental states of collaborators. 
% We leverage the advanced reasoning capabilities of LLMs to predict possible beliefs based on observed information, thereby bridging the gap caused by the lack of collaborative data during pretraining. This model enables agents to reason about the internal states of collaborators and predict the future states of the external world, facilitating more efficient and human-like collaboration.
% Specifically, CoBel-World incorporates two core components. First, inspired by symbolic planning languages such as PDDL~\citep{fox2003pddl2,fabiano2021pddl}, we introduce a symbolic belief language to formalize the multi-agent task settings. Then, the agents will learn knowledge about the external world and represent it as belief rules to guide subsequent task execution through a collaborative propose-and-revise progress. Second, each agent maintains a dynamic internal world model with beliefs. This belief world model is updated via reasoning to infer the intents of collaborators from partial observation and predict the possible states of external world. 
\duo{By leveraging the advanced reasoning capabilities of LLMs, CoBel-World enables agents to reason about the internal states of collaborators and predict the future states of the environment, thus facilitating more efficient and human-like collaboration. Specifically, CoBel-World incorporates two core components. First, inspired by symbolic planning languages such as PDDL~\citep{fox2003pddl2,fabiano2021pddl}, we introduce a symbolic belief representation module to translate natural language descriptions of the open-ended world into symbolic, structured representations of beliefs. Agents then can use these symbolic beliefs to derive belief rules that guide task execution through a collaborative propose-and-revise process. Second, we design a Bayesian belief collaboration protocol that operates in the spirit of Bayesian filter. This protocol harnesses LLM reasoning to predict possible beliefs and detect potential miscoordination in a zero-shot manner, without requiring additional data for LLM fine-tuning. CoBel-World uses this protocol to dynamically update each agent’s belief world model, ensuring consistent multi-agent collaboration even in partially observable environments.}

To summarize, this work makes the following contributions:
\begin{itemize}    
    \item We propose {{CoBel-World}}, a novel framework that integrates a collaborative belief world into LLM agents, enabling efficient communication and consistent planning. 
    \item We design a {symbolic belief representation module} to represent the world knowledge in a structured and explicit form to guide collaboration. We further design a {{Bayesian belief collaboration}} protocol in a Bayesian filter manner, demonstrating how to leverage LLM reasoning capabilities to \duo{predict beliefs and detect miscoordination in a zero-shot manner.}
    \item We evaluate CoBel-World on challenging embodied collaboration \duo{benchmarks TDW-MAT and C-WAH with open-ended environments} \citep{Zhang2023BuildingCE}. Results show that CoBel-World reduces \duo{{the average}} communication cost by  \textbf{\textit{64-79\%}} while improving \duo{{the average}} task completion efficiency by  \textbf{\textit{4-28\%}} compared to  state-of-the-art  methods, demonstrating the efficacy of belief-driven collaboration.
\end{itemize}
%%%%%%%%%%%%%%%%%%%%%%%%%%%%%%%%%%%%%%%

%%%%%%%%%%%%%%%%%%%%%%%%%%%%%%%%%%%%%%%

%%%%%%%%%%%%%%%%%%%%%%%%%%%%%%%%%%%%%%%

%%%%%%%%%%%%%%%%%%%%%%%%%%%%%%%%%%%%%%%
\begin{figure*}[!t]
    \centering
    \includegraphics[width=0.96\linewidth]{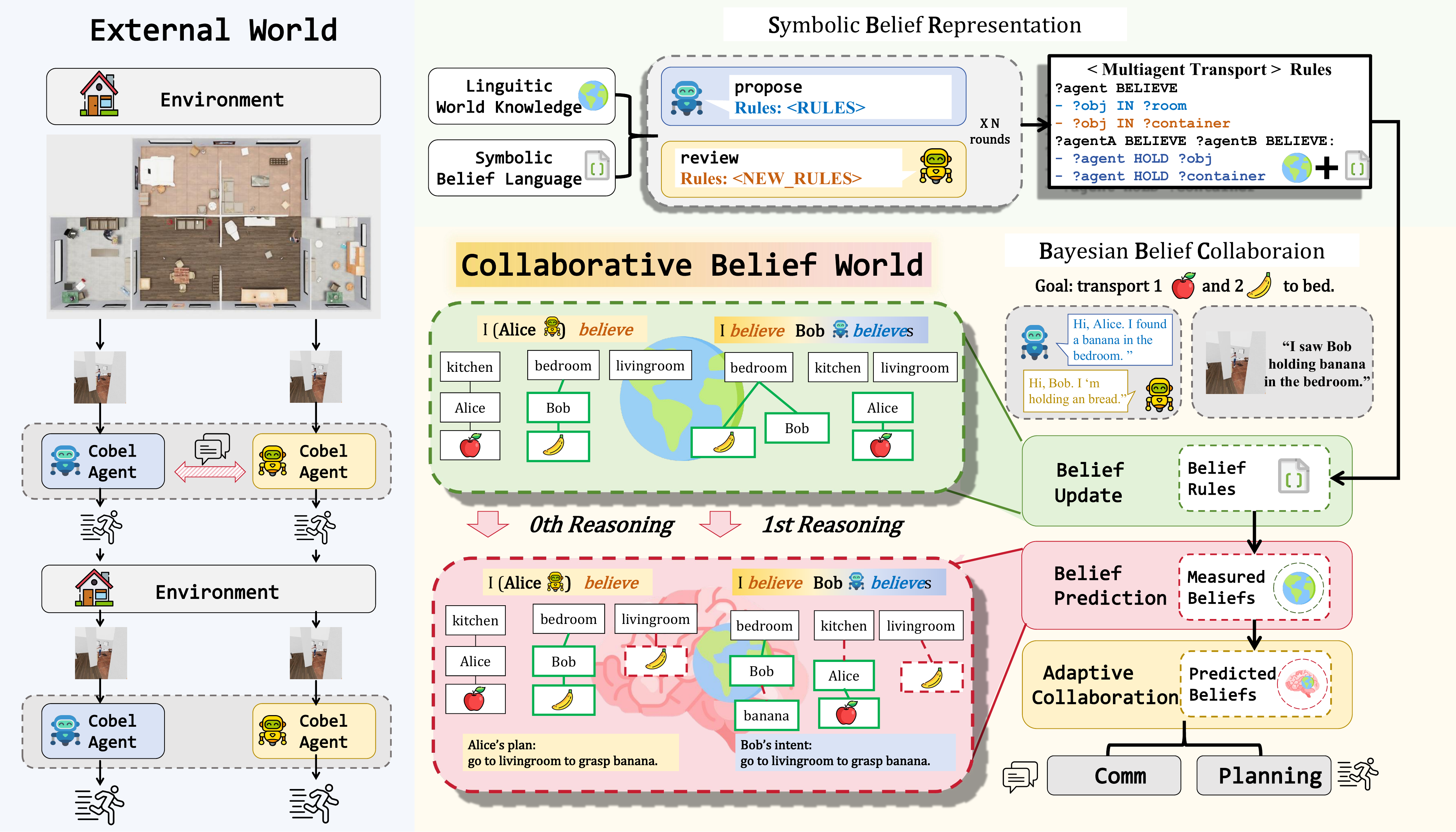}
    \vspace{-0.1cm}
    \caption{\textbf{Overview of CoBel-World.}
Cobel-World comprises two key components: (1) \textbf{Symbolic belief representation}: All agents are organized in a collaborative reasoning process to analyze the requirements of the task and summarize the rules in a structured format. The resulting consensus set of belief rules forms the collaborative belief world. (2) \textbf{Bayesian belief collaboration}: After the belief world is constructed, each agent updates it through \textbf{belief update} and \textbf{belief prediction}, both of which are facilitated by LLM reasoning. Adaptive collaborative decisions will be made based on the beliefs. }
    \vspace{-0.2cm}
    \label{fig:framework}
\end{figure*}
%%%%%%%%%%%%%%%%%%%%%%%%%%%%%%%%%%%%%%%

%%%%%%%%%%%%%%%%%%%%%%%%%%%%%%%%%%%%%%%
\section{Related Work}
\label{related_works}
% \textbf{LLM-based multi-agent collaboration and communication.}
\duo{\textbf{LLM-based multi-agent collaboration.}}
% Recent advances in large language models (LLMs) have enabled their deployment as autonomous agents capable of reasoning, planning, and communication in collaborative settings. Systems such as MetaGPT~\citep{hong2023metagpt} and ChatDev~\citep{qian2023chatdev} demonstrate that LLM agents can follow predefined workflows to solve complex tasks. 
\duo{Recent efforts such as MetaGPT \citep{hong2023metagpt} and ChatDev \citep{qian2024chatdev} have explored the use of LLMs for collaborative task solving.}
% In embodied intelligence, frameworks like CoELA~\citep{Zhang2023BuildingCE}, CaPo~\citep{Liu2024CaPoCP}, and RoCo~\citep{mandi2024roco} integrate LLMs with perception and action modules to support collaborative embodied tasks. 
In particular, CoELA~\citep{Zhang2023BuildingCE}, CaPo~\citep{Liu2024CaPoCP}, and RoCo~\citep{mandi2024roco} integrate LLMs with perception and action modules to support collaborative embodied tasks \duo{in open-ended environments.}
However, these approaches typically rely on fixed communication protocols, such as tep-by-step message generation~\citep{Zhang2023BuildingCE}, event-driven multi-round discussion~\citep{Liu2024CaPoCP}, or dense discussion~\citep{guo2024embodied}, leading to excessive communication overhead and poor scalability under partial observability.
% \re{Recently, several benchmarks have been developed to evaluate LLM-based multi-agent systems in embodied environments. PARTNR~\citep{Chang2024PARTNRAB} provides a large-scale suite of household tasks to evaluate the reasoning and planning capabilities of LLM-based multi-agent systems. CoELA~\citep{Zhang2023BuildingCE} introduces the TDW-MAT and CWAH benchmarks, offering embodied multi-agent tasks that require agents to communicate explicitly to complete complex tasks. Other works, such as CaPo~\citep{Liu2024CaPoCP} and RoCo~\citep{mandi2024roco}, further investigate diverse collaboration frameworks.} However, these approaches typically rely on fixed communication protocols, such as \re{step}-by-step message generation~\citep{Zhang2023BuildingCE}, event-driven multi-round discussion~\citep{Liu2024CaPoCP}, or dense discussion~\citep{guo2024embodied}, leading to excessive communication overhead and poor scalability under partial observability. 
In contrast, our work introduces a belief-driven communication mechanism that enables LLM agents to dynamically identify and exchange only the most valuable information, significantly reducing communication redundancy while improving collaboration efficiency.

% \noindent\textbf{Belief modeling in decentralized multi-agent systems.}
\noindent\textbf{Belief modeling in multi-agent systems.}
In decentralized partially observable Markov decision process (DEC-POMDP), belief modeling is central to enabling agents to maintain and update probabilistic estimates over hidden states and other agents’ intents~\citep{kominis2015beliefs,moreno2021neural}. Techniques such as Bayesian reasoning~\citep{foerster2019bayesian} and probabilistic recursive reasoning~\citep{wen2019probabilistic} allow agents to infer unobserved variables and align internal states through belief estimation. More recent approaches leverage pretrained belief models~\citep{zhai2023dynamic,pritz2025belief}, achieving improved collaboration in cooperative games such as Hanabi and Overcooked. \re{\citet{wu2020too} leverages inverse planning to infer collaborators’ beliefs, allowing agents to dynamically switch between task division and joint collaboration.} \re{~\citet{jha2024neural} enables agents to perform higher-order belief modeling with significantly reduced computational cost.} \re{~\citet{cao2024enhancing} incorporates logical rules to infer human goals and beliefs from demonstrations, thereby guiding hierarchical human–AI collaboration.} While promising, these methods are largely limited to low-dimensional, discrete-state environments with handcrafted features or require extensive training data. Our work bridges this gap by leveraging the zero-shot reasoning capabilities of LLMs to construct and update structured belief representations in high-dimensional, open-ended physical environments without environment-specific training.

\re{Recent works~\citep{yi2025debate,Zhang2024COMBOCW} attempt to incorporate belief modeling into LLM-based multi-agent systems to guide decision and strategy selection. However, these works primarily operate under communication-free settings, which limits their scalability in real-world partially observable environments. In contrast, CoBel-World leverages structured belief modeling to guide communication behaviors. Agents with such collaborative belief world can proactively determine when, who and how to communicate.}

% \noindent\textbf{Reasoning capabilities of LLM-based agents.}
% The effectiveness of LLMs as autonomous agents hinges on their ability to perform diverse forms of reasoning, from task planning to social inference. Recent work has demonstrated that structured reasoning paradigms significantly enhance agent performance in complex tasks. Notable works include Chain-of-Thought (CoT)~\citep{wei2022chain} and Tree of Thoughts (ToT)~\citep{yao2023tree}, which introduces multi-step reasoning to solve complex problems. More recently, research has advanced social reasoning, particularly theory of mind (ToM), enabling agents to model others’ beliefs, intents, and internal states~\citep{he2023hi,sclar-etal-2023-minding,Jin2024MMToMQAMT,Shi2024MuMAToMMM}. Several works~\citep{Li2023TheoryOM,ma2023towards,zhang2025autotom}, have gained benefits in collaborative multi-agent tasks with the introduction of such ability. 

% \re{AutoToM~\citep{zhang2025autotom} automates model-based ToM to robustly infer latent mental variables across domains.} 
%最近，许多工作为LLM-Based multi-agent system 提供了可测试的benchmark. PARTNR提供了大量的任务集和完善的测试和评估基础设施，同时给出领域讨论和insight. CoELA则提供了通信式多智能体的测试基准TDW-MAT和CWAH. 其他工作探索了不同的协作框架（coela,capo,roco）
%In embodied intelligence, frameworks like CoELA~\citep{Zhang2023BuildingCE}, CaPo~\citep{Liu2024CaPoCP}, and RoCo~\citep{mandi2024roco} integrate LLMs with perception and action modules to support collaborative embodied tasks.
%%%%%%%%%%%%%%%%%%%%%%%%%%%%%%%%%%%%%%%

%%%%%%%%%%%%%%%%%%%%%%%%%%%%%%%%%%%%%%%
\section{Method}
\label{method}
% In this section, we present CoBel-World leverages belief modeling to address communication redundancy and inconsistent collaboration in embodied multi-agent systems. The theoretical foundation of CoBel-World can be found in Appendix~\ref{appendix:formulation}. Following the paradigm of belief modeling in traditional MARL, we decompose the construction of CoBel-World framework into two components: \textbf{Symbolic Belief Representation} for belief representation and \textbf{Bayesian Belief Collaboration} for belief update, as depicted in Figure~\ref{fig:framework}.

In this section, we present \textbf{\textit{CoBel-World}}, a principled framework leveraging belief modeling to mitigate communication redundancy and collaborative misalignment in multi-agent systems. Following the paradigm of belief modeling in traditional MARL, we decompose CoBel-World into two seamlessly integrated components: \textbf{symbolic belief representation} (detailed in $\S$\ref{sec:method-2}) for belief construction and \textbf{Bayesian belief collaboration} (detailed in $\S$\ref{sec:method-3}) for belief update. The overall framework of CoBel-World is depicted in Figure \ref{fig:framework}.

% \subsection{overall framework}
% \label{sec:method-1}
% First, Symbolic Belief Representation (detailed in Section~\ref{subsec:symbolic_belief_repre}), centered on a symbolic belief language, enables agents to autonomously interpret task requirements in open-ended environments and encode world knowledge into structured belief rules. It further incorporates a collaborative reasoning process to establish a \textit{collaborative belief world}, ensuring consistent modeling of the environment and collaborators' intents. Second, {Bayesian Belief Collaboration} (detailed in Section~\ref{methodbayesianbeliefcollab}) maintains and dynamically updates the established belief world during task execution. Agents perform belief updates via a Bayesian filtering scheme powered by LLM reasoning to detect potential miscoordination. When belief misalignment arise, agents proactively communicate to align beliefs and share intents; when beliefs are synchronized, they proceed with action planning and execution. This adaptive mechanism enables context-aware collaboration decisions based on task progress and collaborators' evolving states.

% 我们基于POMDP中信念更新的数学方程——贝叶斯滤波器将我们的CoBel-World框架分为两部分：
\subsection{Symbolic Belief Representation}
\label{sec:method-2}
% LLMs struggle to accurately model diverse and structured beliefs due to the complexity of real-world environments. To this end, we introduce symbolic belief representation to model beliefs in a strutured and stable form.

\duo{In this section, we delve into the detailed design of symbolic belief representation, which consists of a symbolic belief language for structured belief representations and a collaborative belief initialization scheme for belief construction.}

% \noindent\textbf{Formalizing the belief language.} Inspired by classical planning languages~\citep{fox2003pddl2,fabiano2021pddl}, we formalize beliefs as tuples consisting of entities, attributes, and predicates. In particular, since beliefs are inherently higher-order (e.g., “Bob believes that Alice believes the apple is in the living room”), we explicitly introduce a recursive belief predicate BELIEVE to capture the collaborators' mental states. The definition of \textit{symbolic belief language} is as follows:
\noindent\duo{\textbf{Symbolic belief language.}} Inspired by classical planning languages~\citep{fox2003pddl2,fabiano2021pddl}, we formalize beliefs as tuples consisting of entities, attributes, and predicates. In particular, since beliefs are inherently higher-order (e.g., “Bob believes that Alice believes the apple is in the living room”), we explicitly introduce a recursive belief predicate  to capture the collaborators' mental states. Specifically, we design the symbolic belief language as follows:
% To ensure a compact and interpretable representation, we formalize the state $\phi$ and the belief $\mathcal{B}$ as follows:
\begin{align}
    &\text{\textbf{Atomic state:}} && \nonumber \\
    &\hspace{0.4em} s ::= \langle e_i, \textsc{Pred}, e_j \rangle \mid \langle e_i, \textsc{Prop}, v \rangle, && \\[0.5ex]
    &\text{\textbf{Zero-order belief:}} && \nonumber \\
    &\hspace{0.4em} b^0 ::= n_i \ \textsc{Bel} \ s, && \\[0.5ex]
    &\text{\textbf{First-order belief:}} && \nonumber \\
    &\hspace{0.4em} b^1 ::= n_i \ \textsc{Bel} \ (n_j \ \textsc{Bel} \ s), &&
\end{align}
% {where $e \in \mathcal{E}$ and $\mathcal{E}$ is the set of entities (e.g., \texttt{Alice}, \texttt{apple})}; 
\duo{where entity $e \in \mathcal{E}$ and $\mathcal{E}$ denotes the set of entities including all agents (e.g., \texttt{Alice}) and objects (e.g., \texttt{apple})}; 
$\hlgray{\textsc{Pred}}$ represents relational descriptors (e.g. \texttt{In}, \texttt{Hold}); $\hlgray{\textsc{Prop}}$ denotes entity properties (e.g., exploration status); and $v \in \mathcal{V}$ defines the discrete values (e.g., \texttt{part}, \texttt{all}). The operator $\hlgray{\textsc{Bel}}$ serves as a connective representing an agent's mental state, mapping an agent $n_i \in \mathcal{N}$ to a state $s$ or another belief $b$. Here, $\mathcal{N}$ denotes the set of all agents, with $n_i, n_j$ representing specific instances.

\begin{figure}[t]
    \centering
    \includegraphics[width=0.98\linewidth]{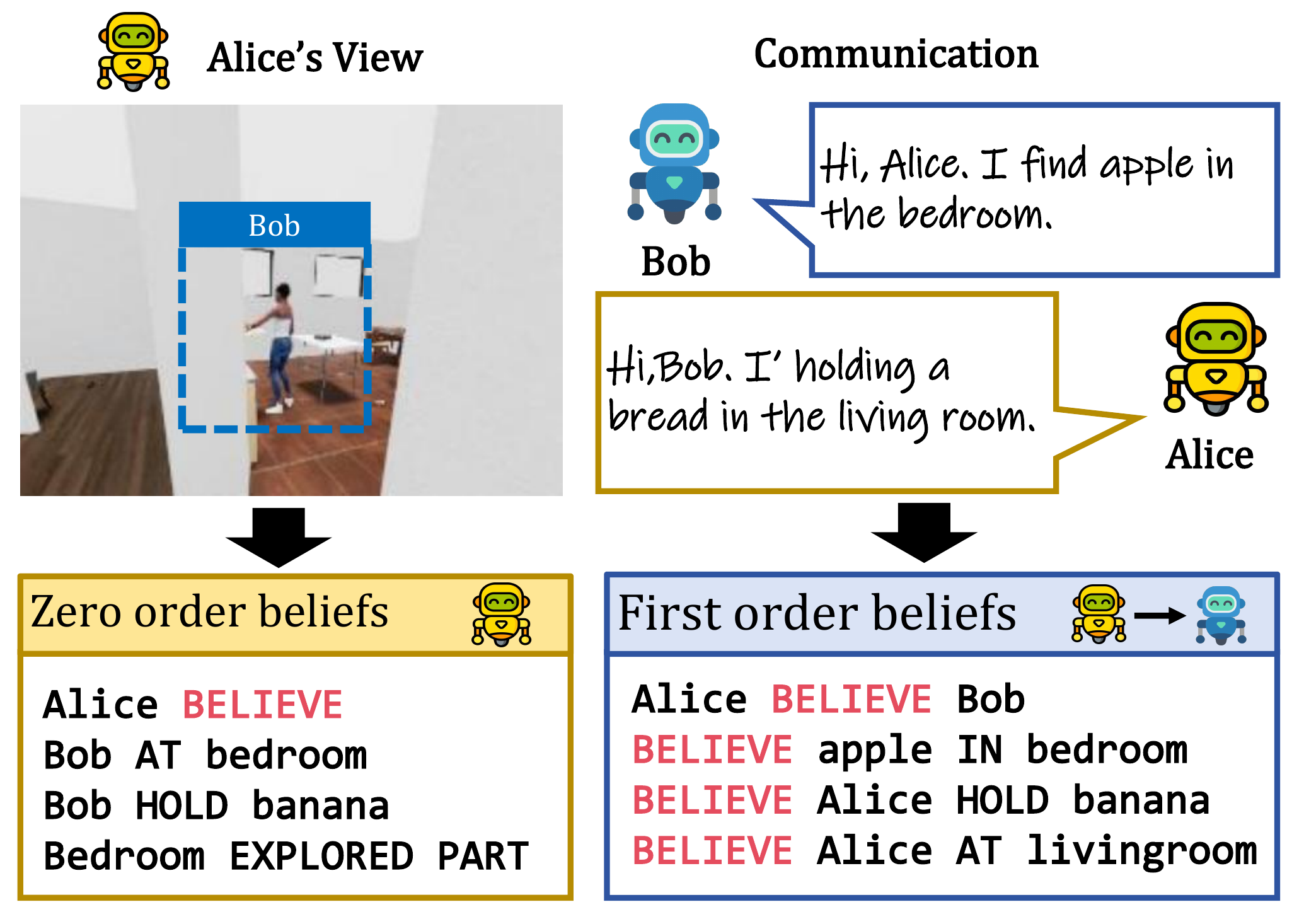}
    \vspace{-0.3cm}
    \caption{Examples of the transformation from unstructured \duo{observations} to structured beliefs.}
    \label{fig:belief-representation}
\end{figure}

\duo{Figure~\ref{fig:belief-representation} provides a concrete example for our belief representations.} When agent Alice observes the other agent Bob holding a banana, Alice constructs a zero-order belief as: \hl{\texttt{Alice \,BELIEVE \,Bob \,HOLD \,banana}}. When Alice receives the message “I found an apple in the bedroom” from Bob, this information is interpreted as a first-order belief: \hl{\texttt{Alice \,BELIEVE \,Bob \,BELIEVE \,apple \,IN \,bedroom}}.

% \noindent\textbf{Collaborative Belief Initialization.} As shown in Figure~\ref{fig:framework}, we propose a \textit{Collaborative Propose-and-Revise Protocol} to ensure all agents collaborate on a consistent belief recognition. Instead of relying on a single agent's potentially hallucinated interpretation, agents undergo a peer-review reasoning loop to formalize \textit{Belief Rules} $\mathcal{R}$. These rules serve as the foundation for the subsequent Bayesian update process.

\noindent\textbf{Collaborative belief initialization.} \duo{Solely relying on the agent itself to construct beliefs is prone to hallucination issues. Hence, we introduce a collaborative belief initialization mechanism where all agents collaborate on belief construction. Specifically, rather than constructing beliefs independently, agents jointly refine a shared set of symbolic belief rules through iterative proposal and cross-agent review. These rules encode task constraints, agent capabilities, and logical dependencies. The resulting consensus set of belief  rules forms the collaborative belief world that lays the  foundation for the subsequent Bayesian update process.}

% In this progress, agents iteratively propose and revise the structured belief rules including task constraints, agent capabilities, and logical dependencies. The output of this collaborative progress is a consensus set of belief rules, which constitute a common collaborative belief world and are then used to guide subsequent task execution. 

\subsection{Bayesian Belief Collaboration}
\label{sec:method-3}
In DEC-POMDP, belief modeling follows a Bayesian filter~\citep{chen2003bayesian} paradigm (detailed in Appendix \ref{appendix:formulation}): a \textbf{update} step that incorporates posterior observation, followed by a \textbf{prediction} step based on prior beliefs. We adopt this well-grounded mathematical structure \duo{for CoBel-World}. 
% In the update phase, we generate the agent’s beliefs $\mathcal{B}_t$ using its partial observation $o_t$ from the environment. 
\duo{In the update phase, the agent updates its beliefs in response to partial observations.}
In the prediction phase, we leverage the reasoning capabilities of LLMs to predict the potential states of the external environment and infer collaborator’s intents. The design details are elaborated as follows.

\noindent\textbf{Belief update.} 
% This step captures the agent’s ability to update its beliefs in response to partial observation. 
\duo{Given the belief world which consists of a set of belief rules $\mathcal{R}$ constructed in the first phase, the agent  updates two levels of beliefs via LLM reasoning.}%both \textbf{zero-order beliefs} and \textbf{first-order beliefs}. 
% Notably, during the update of first-order beliefs, we employ theory-of-mind (ToM) reasoning~\citep{Li2023TheoryOM,ma2023towards} to prompt the agent to interpret messages from the collaborator’s perspective. This prevents the agent from \duo{confusing} personal information with public information, ensuring a more accurate belief estimation. 
% This process is formulized below:
\begin{align}
    b_t^0 &= LLM_{\text{update\_zero}}(\mathcal{R}, o_t), \label{eq:update_zero} \\
    b_t^1 &= LLM_{\text{update\_first}}(\mathcal{R}, o_c), \label{eq:update_first}
\end{align}
% where $\mathcal{R}$ is the belief rules. 
where $o_t = (o_v, o_c)$ is the partial observation acquired from the environment \duo{and collaborators, $o_v$ represents the ego-centric visual perception (e.g., object positions), and $o_c$ denotes the communication message explicitly transmitted by other agents. }\duo{Notably, during the update of first-order beliefs, we employ theory-of-mind (ToM) reasoning~\citep{Li2023TheoryOM,ma2023towards,Strachan2024TestingTO} to prompt the agent to interpret messages from the collaborator’s perspective. This prevents the agent from \duo{confusing} personal information with public information, ensuring a more accurate belief estimation~\citep{Shi2024MuMAToMMM,zhang2025autotom}. }
% The observation $o_t$ can be decomposed into two modalities: \textbf{$o_v$}: The ego visual perception. (e.g., object positions, agent states); \textbf{$o_c$}: The communication contents explicitly transmitted by other agents.

%%he2023hi,sclar-etal-2023-minding,Jin2024MMToMQAMT,Shi2024MuMAToMMM

\noindent \textbf{Belief prediction. } 
% Building upon the agent’s collaborative belief world, we enable proactive coordination by predicting the possible beliefs based on the updated beliefs. Agents perform belief prediction separately based on zero-order and first-order beliefs. For zero-order beliefs, we prompt the LLM to infer possible states of environment. 
% Based on these predicted beliefs, agents then generate plans that maximize task efficiency by prioritizing high-utility, low-uncertainty exploration or manipulation steps. 
\duo{Based on the updated beliefs, the agent proactively predicts the possible beliefs in the future to maintain the knowledge about the external environment and the intents of collaborators. Specifically, the agent predicts the future zero-order belief to reason the possible states of environment~\citep{wei2022chain}. Based on these predicted beliefs, agents then generate plans that maximize task efficiency by prioritizing high-utility, low-uncertainty exploration or manipulation steps. This process is formally defined as:}
\begin{align}
    {b}_{t+1}^0 &= LLM_{\text{reason}}({b}_t^0, G, P), 
    \label{predict_zero} \\
    \pi_{t+1} &= LLM_{\text{plan}}(\bar{b}_{t+1}^0, G, P), 
\end{align}
\duo{where ${b}_{t+1}^0$ represents the predicted zero-order beliefs, $\pi_{t+1}$ represents the plan the agent will take, $G$ is the task goal and $P$ is the description of task progress which are both provided by the environment.}

% For first-order beliefs, we repeat the reasoning step. However, to ensure comprehensive coverage of potential miscoordination, agent will explicitly reasons over multiple intents for every collaborator—not just the most likely one. This multi-hypothesis modeling allows the agents to fully assess the current collaboration status, guiding their subsequent collaboration behaviors. This process is formulized below:

\duo{Next, the agent predicts the first-order beliefs to infer the intents of collaborators and then reasons about  their possible plans to avoid conflicts and miscoordination.}
% \begin{tcolorbox}
% [colback=Emerald!10,colframe=cyan!40!black,title=\textbf{Belief Prediction Process}]
\begin{align}
    % \bar{\mathcal{B}}_{t+1}^0 &= f_{\text{reason}}(\bar{\mathcal{B}}_t^0, G, P) 
    % \label{predict_zero} \\
    % \pi_{t+1}^{A_i} &= f_{\text{plan}}(\bar{B}_{t+1}^0, G, P) \\
    {b}_{t+1}^1 &= LLM_{\text{reason}}({b}_t^1, G, P), \\
    \bar\pi_{t+1} &= LLM_{\text{plan}}({b}_{t+1}^1, G, P),
\end{align}
\duo{where ${b}_{t+1}^1$ represents the predicted first-order beliefs, $\bar\pi_{t+1}$ represents the predicted possible plans of collaborators. }
% \end{tcolorbox}
% The prompt structure is illustrated below:

% \begin{tcolorbox}
% [colback=Emerald!10,colframe=cyan!40!black,title=\textbf{Belief Prediction Prompt}]
% First-order Belief Prediction: \texttt{<Instruct Head>} + \texttt{<first-order Beliefs>}

% LLM: \texttt{<Predicted Beliefs>} + \texttt{<Collaborator's Intents>}

% Zero-order Belief Prediction: \texttt{<Instruct Head>} + \texttt{<zero-order Beliefs>}

% LLM: \texttt{<Predicted Beliefs>} + \texttt{<My plan>}
% \end{tcolorbox}

\noindent\textbf{Adaptive collaboration. }
% After updating and predicting the collaborative belief world, each agent obtains an estimation about collaborators' intents and their mental states, enabling agents to proactively evaluate the current collaboration status. With this capability for dynamic intent inference and state estimation, agents can autonomously and adaptively decide how to collaborate: when potential miscoordination (e.g. conflicting plans) is detected, they send context-aware messages to promote consensus and consistent planning among collaborators; when the current collaboration status is unlikely to cause serious conflicts, agents prefer executing actions directly to improve overall efficiency. 
\duo{After updating and predicting the collaborative belief world, each agent obtains an estimation about the external world, and thus can adaptively decide the collaboration strategies. For example, when potential miscoordination (e.g. conflicting plans) is detected, they send context-aware messages to promote consensus and consistent planning among collaborators. In contrast, when serious conflicts are unlikely to occur, agents can prefer directly executing actions to improve overall efficiency.} 

\duo{To be specific, we implement the adaptive collaboration scheme with two steps. First, we prompt the agents to explicitly reason over two key aspects:}
% To be specific, we first prompt the LLM to explicitly reason over two key aspects: 
i) belief misalignment (e.g., only Bob knows the apple’s location), and ii) potentially conflicting actions (e.g., Alice and Bob plan to explore the same room). Second, if agents detect the potential miscoordination, they construct a message with the misaligned beliefs and share their intents. Based on this reasoning analysis, agents autonomously adjust their collaboration strategies, thus achieving efficient, adaptive, and intent-aware collaboration. Details are illustrated in  Figure~\ref{fig:framework}.

%%%%%%%%%%%%%%%%%%%%%%%%%%%%%%%%%%%%%%%

%%%%%%%%%%%%%%%%%%%%%%%%%%%%%%%%%%%%%%%
\section{Experienment}
\label{experiement}
%%%%%%%%%%%%%%%%%%%%%%%%%%%%%%
\definecolor{rcolor}{RGB}{234,235,255}
\definecolor{lightpink}{RGB}{255,224,224}
\setlength{\fboxsep}{0pt}

\begin{table*}[t]
\small
\centering
\renewcommand{\arraystretch}{1.3} 
\setlength{\tabcolsep}{3pt}      
\newcommand{\std}[2]{#1$_{{\scriptscriptstyle \pm#2}}$}

\caption{Performance comparison on TDW\_MAT benchmark. ``$\uparrow$/$\downarrow$'' means higher/lower is better. 
%Results are highlighted per LLM group: \colorbox{lightpink}{pink} for the best and \underline{underlined} for the second-best performance within each group. Subscripts $\pm$ denote the standard deviation over multiple independent trials.
\duo{Results in \colorbox{lightpink}{pink}  and \underline{underlined} denote the best and second-best performance in each LLM group, respectively, Subscripts $\pm$ indicate standard deviation across three independent trials.}
}
\vspace{-0.25cm}
\resizebox{\textwidth}{!}{%
\begin{tabular}{C{2.5cm}|cc|ccc|ccc|ccc}
\toprule
 \multirow{2}{*}{\textbf{Task Category}} & \multicolumn{2}{c|}{\textbf{Classic Agents}} & \multicolumn{3}{c|}{\textbf{Qwen3-32B Agents}} & \multicolumn{3}{c|}{\textbf{DeepseekV3.2 Agents}} & \multicolumn{3}{c}{\textbf{GPT-4o Agents}} \\
\cmidrule(lr){2-3} \cmidrule(lr){4-6} \cmidrule(lr){7-9} \cmidrule(lr){10-12}
 & RHP & RHP+RHP & CoELA & CaPo & \textbf{CoBel-World} & CoELA & CaPo & \textbf{CoBel-World} & CoELA & CaPo & \textbf{CoBel-World} \\
\midrule
\multicolumn{12}{l}{\textit{Transport Rate ($\uparrow$)}} \\
\midrule
Food-low-cap & \std{50.00}{0.00} & \std{75.00}{0.00} & \underline{\std{73.89}{3.47}} & \std{62.22}{1.92} & \cellcolor{lightpink}\textbf{\std{84.44}{4.19}} & \underline{\std{61.67}{2.89}} & \std{58.89}{2.51} & \cellcolor{lightpink}\textbf{\std{86.11}{0.96}} & \underline{\std{79.44}{3.47}} & \std{79.44}{4.20} & \cellcolor{lightpink}\textbf{\std{87.22}{4.20}} \\

Stuff-low-cap & \std{50.00}{0.00} & \std{75.00}{0.00} & \underline{\std{73.89}{4.19}} & \std{58.33}{4.41} & \cellcolor{lightpink}\textbf{\std{85.00}{0.00}} & \underline{\std{59.44}{5.85}} & \std{50.56}{5.09} & \cellcolor{lightpink}\textbf{\std{76.11}{1.92}} & \underline{\std{77.22}{2.55}} & \std{75.56}{7.88} & \cellcolor{lightpink}\textbf{\std{80.00}{0.00}} \\

\textbf{Low-cap Avg} & \std{50.00}{0.00} & \std{75.00}{0.00} & \underline{\std{73.89}{3.47}} & \std{60.28}{3.15} & \cellcolor{lightpink}\textbf{\std{84.72}{2.10}} & \underline{\std{60.56}{1.92}} & \std{54.72}{1.27} & \cellcolor{lightpink}\textbf{\std{81.11}{0.48}} & \underline{\std{78.33}{1.44}} & \std{77.50}{4.64} & \cellcolor{lightpink}\textbf{\std{83.61}{2.10}} \\
\midrule
Food-high-cap & \std{40.00}{0.00} & \std{73.33}{0.00} & \underline{\std{75.00}{8.66}} & \std{67.22}{9.62} & \cellcolor{lightpink}\textbf{\std{86.11}{8.22}} & \underline{\std{63.89}{5.85}} & \std{59.44}{0.96} & \cellcolor{lightpink}\textbf{\std{75.56}{0.96}} & \underline{\std{86.67}{1.67}} & \std{82.22}{5.09} & \cellcolor{lightpink}\textbf{\std{87.78}{1.92}} \\

Stuff-high-cap & \std{46.67}{0.00} & \std{80.00}{0.00} & \cellcolor{lightpink}\textbf{\std{86.11}{3.85}} & \std{70.56}{15.12} & \underline{\std{82.78}{5.36}} & \underline{\std{66.67}{2.89}} & \std{59.44}{6.74} & \cellcolor{lightpink}\textbf{\std{78.33}{0.00}} & \std{79.44}{5.36} & \underline{\std{80.56}{1.92}} & \cellcolor{lightpink}\textbf{\std{85.56}{1.92}} \\

\textbf{High-cap Avg} & \std{43.33}{0.00} & \std{76.67}{0.00} & \underline{\std{80.56}{5.55}} & \std{68.89}{2.93} & \cellcolor{lightpink}\textbf{\std{84.44}{4.28}} & \underline{\std{65.28}{1.73}} & \std{59.44}{2.93} & \cellcolor{lightpink}\textbf{\std{76.94}{0.48}} & \underline{\std{83.06}{2.68}} & \std{81.39}{2.55} & \cellcolor{lightpink}\textbf{\std{86.67}{1.67}} \\
\midrule
\rowcolor{rcolor}
\textbf{Total Average} & \std{46.67}{0.00} & \std{75.83}{0.00} & \underline{\std{77.22}{1.05}} & \std{64.58}{3.00} & \cellcolor{lightpink}\textbf{\std{84.58}{1.10}} & \underline{\std{62.92}{1.10}} & \std{57.08}{1.44} & \cellcolor{lightpink}\textbf{\std{79.72}{1.05}} & \underline{\std{80.69}{1.05}} & \std{79.44}{2.06} & \cellcolor{lightpink}\textbf{\std{85.14}{1.05}} \\
\midrule
\multicolumn{12}{l}{\textit{Communication Cost ($\downarrow$)}} \\
\midrule
Food-low-cap  & --- & --- & \underline{\std{3921.33}{384.22}} & \std{7936.28}{924.44} & \cellcolor{lightpink}\textbf{\std{1537.00}{124.25}} & \underline{\std{3174.67}{595.31}} & \std{6902.06}{779.59} & \cellcolor{lightpink}\textbf{\std{616.22}{75.44}} & \underline{\std{2569.83}{196.92}} & \std{4572.03}{932.33} & \cellcolor{lightpink}\textbf{\std{937.44}{75.13}} \\

Stuff-low-cap & --- & --- & \underline{\std{3304.44}{147.34}} & \std{7424.33}{703.84} & \cellcolor{lightpink}\textbf{\std{1354.06}{198.50}} & \underline{\std{2901.06}{138.85}} & \std{6832.39}{1297.08} & \cellcolor{lightpink}\textbf{\std{809.67}{232.38}} & \underline{\std{2586.44}{149.51}} & \std{3950.36}{844.17} & \cellcolor{lightpink}\textbf{\std{833.11}{15.42}} \\

\textbf{Low-cap Avg} & --- & --- & \underline{\std{3612.89}{238.95}} & \std{7680.31}{466.14} & \cellcolor{lightpink}\textbf{\std{1445.53}{88.19}} & \underline{\std{3037.86}{332.14}} & \std{6867.22}{466.80} & \cellcolor{lightpink}\textbf{\std{712.94}{153.91}} & \underline{\std{2578.14}{173.15}} & \std{4261.69}{888.25} & \cellcolor{lightpink}\textbf{\std{885.28}{43.62}} \\
\midrule
Food-high-cap & --- & --- & \underline{\std{3895.78}{189.59}} & \std{7786.89}{800.99} & \cellcolor{lightpink}\textbf{\std{1579.39}{330.99}} & \underline{\std{3331.56}{884.38}} & \std{7142.50}{323.73} & \cellcolor{lightpink}\textbf{\std{510.67}{16.17}} & \underline{\std{2523.83}{118.60}} & \std{4673.69}{1140.50} & \cellcolor{lightpink}\textbf{\std{937.67}{178.62}} \\

Stuff-high-cap & --- & --- & \underline{\std{3817.33}{86.31}} & \std{7697.72}{875.93} & \cellcolor{lightpink}\textbf{\std{1430.44}{181.53}} & \underline{\std{2593.67}{20.89}} & \std{7330.17}{649.90} & \cellcolor{lightpink}\textbf{\std{627.17}{24.83}} & \underline{\std{2366.89}{249.92}} & \std{4529.42}{1065.67} & \cellcolor{lightpink}\textbf{\std{894.83}{148.18}} \\

\textbf{High-cap Avg} & --- & --- & \underline{\std{3856.56}{52.39}} & \std{7742.31}{807.86} & \cellcolor{lightpink}\textbf{\std{1504.92}{94.08}} & \underline{\std{2962.61}{446.63}} & \std{7236.33}{451.23} & \cellcolor{lightpink}\textbf{\std{568.92}{4.33}} & \underline{\std{2445.36}{175.10}} & \std{4602.06}{1103.08} & \cellcolor{lightpink}\textbf{\std{916.25}{162.25}} \\
\midrule
\rowcolor{rcolor}
\textbf{Total Average} & --- & --- & \underline{\std{3734.72}{112.63}} & \std{7711.31}{182.13} & \cellcolor{lightpink}\textbf{\std{1475.22}{35.22}} & \underline{\std{3000.24}{196.27}} & \std{7051.78}{158.52} & \cellcolor{lightpink}\textbf{\std{640.93}{74.79}} & \underline{\std{2511.75}{151.99}} & \std{4432.71}{995.67} & \cellcolor{lightpink}\textbf{\std{900.76}{86.77}} \\
\bottomrule
\end{tabular}
}
\label{tab:tdw_mat_updated_classic}
\end{table*}
%%%%%%%%%%%%%%%%%%

% In this section, we instantiate CoBel-World with diverse LLMs to validate its effectiveness across different benchmarks. First, we compare CoBel-World against several important baselines to demonstrate its superiority in both collaboration efficiency and communication cost. 
% Second, we visualize task trajectories and interaction content to illustrate how CoBel-World leverages belief modeling to facilitate consistent planning and effective communication. 
% Next, we conduct ablation studies to verify the effectiveness of individual modules and extend CoBel-World to scenarios involving more agents to validate its scalability in many-agent environments.

\subsection{Setup}
\label{experienment-settings}

\noindent\textbf{Benchmarks.} \duo{Recent efforts have established several benchmarks to evaluate LLM-based multi-agent systems in open-ended environments~\citep{Chang2024PARTNRAB,Zhang2023BuildingCE}. To demonstrate CoBel-World's efficiency in communication, we follow CoELA~\citep{Zhang2023BuildingCE} and adopt the two challenging embodied multi-agent benchmarks for our experiments: TDW-MAT~\citep{Zhang2023BuildingCE}, and the C-WAH~\citep{Zhang2023BuildingCE}. TDW-MAT is built on the general purpose virtual world simulation platform TDW~\citep{gan2020threedworld}, and requires agents to move objects to the destination by their hands or containers. 
% Moreover, agents can receive ego-centric RGB-D images as observation and communicate with others. 
In C-WAH, agents are requested to complete five types of household tasks, represented as various predicates with specific counts that must be satisfied. \zm{By default, we use two agents for collaboration.}
More details about TDW-MAT and C-WAH environments are provided in Appendix \ref{tdw} and \ref{cwah}, respectively.}

% \textbf{Metrics.} The evaluation metrics cover two dimension: \textit{task completion efficiency} and \textit{communication cost}. First, task completion efficiency metrics. On TDW-MAT, we adopt Transport Rate, i.e., the fraction of subtasks completed within 3000 time steps (a.k.a. frames), as performance metric. Note, one action step may last multiple time steps, e.g., resetting arms. On C-WAH, Average Steps to complete all tasks is used as the metric to evaluate collaboration efficiency. Second, communication cost, computed by the average number of tokens generated per episode by all agents for generating communication content.

\noindent\textbf{Metrics.} Our evaluation metrics span two dimensions: task completion efficiency and communication cost. For task completion efficiency, we use different metrics for the two benchmarks. On TDW-MAT, we adopt \textit{transport rates} as the primary performance metric, which refers to the fraction of subtasks successfully completed within 3,000 time steps (frames). Note that a single action step may span multiple time steps (e.g., arm resetting). On C-WAH, we report the \textit{average steps} required to complete all tasks, which reflects the efficiency of collaborative coordination. For communication cost, we compute \textit{the average number of tokens} generated by all agents per episode for communication. Higher transport rates, fewer average steps, and fewer tokens indicate better performance.

\definecolor{rcolor}{RGB}{234,235,255}
\definecolor{lightpink}{RGB}{255,224,224}
\setlength{\fboxsep}{0pt}

\begin{table*}[t]
\footnotesize % 基础字号改用 footnotesize
\centering
\renewcommand{\arraystretch}{1.25} % 增加行高，视觉更清晰
\setlength{\tabcolsep}{2.5pt}      % 【关键】极力压缩列间距，为文字换取放大空间
\newcommand{\std}[2]{#1$_{{\scriptscriptstyle \pm#2}}$}

\caption{Performance comparison on C-WAH benchmark. ``$\uparrow$/$\downarrow$'' means higher/lower is better. 
% Pink/underline denote best/second-best per LLM group. Subscripts denote SD.
\duo{Results in \colorbox{lightpink}{pink}  and \underline{underlined} denote the best and second-best performance in each LLM group, respectively, Subscripts $\pm$ indicate standard deviation across three independent trials.}
}
\vspace{-0.25cm}
\resizebox{\textwidth}{!}{%
% 将首列宽度从 2cm 压缩至 1.6cm
\begin{tabular}{C{1.6cm}|c|cc|ccc|ccc|ccc} 
\toprule
 \multicolumn{2}{c|}{\multirow{2}{*}{\textbf{Task / Obs.}}} & \multicolumn{2}{c|}{\textbf{Classic Agents}} & \multicolumn{3}{c|}{\textbf{Qwen3-32B Agents}} & \multicolumn{3}{c|}{\textbf{DeepseekV3.2 Agents}} & \multicolumn{3}{c}{\textbf{GPT-4o Agents}} \\
\cmidrule(lr){3-4} \cmidrule(lr){5-7} \cmidrule(lr){8-10} \cmidrule(lr){11-13}

\multicolumn{2}{c|}{} & MHP & MHP+MHP & CoELA & CaPo & \textbf{CoBel-World} & CoELA & CaPo & \textbf{CoBel-World} & CoELA & CaPo & \textbf{CoBel-World} \\
\midrule

\multicolumn{13}{l}{\textit{Average Step ($\downarrow$)}} \\
\midrule
Prepare & Sym. & \std{155.67}{7.37} & \std{94.00}{2.50} & \cellcolor{lightpink}\textbf{\std{87.17}{13.66}} & \std{98.50}{2.18} & \underline{\std{94.00}{8.67}} & \std{84.83}{8.39} & \underline{\std{72.67}{1.04}} & \cellcolor{lightpink}\textbf{\std{62.33}{7.09}} & \underline{\std{79.50}{6.08}} & \std{85.00}{3.50} & \cellcolor{lightpink}\textbf{\std{64.50}{10.40}} \\
tea & Vis. & \std{211.50}{11.17} & \std{121.50}{8.41} & \underline{\std{169.67}{15.12}} & \std{181.17}{18.66} & \cellcolor{lightpink}\textbf{\std{114.67}{21.83}} & \std{167.17}{27.97} & \underline{\std{114.17}{27.81}} & \cellcolor{lightpink}\textbf{\std{76.83}{4.95}} & \std{135.00}{10.44} & \underline{\std{119.00}{0.87}} & \cellcolor{lightpink}\textbf{\std{54.00}{4.09}} \\
\hline
Wash & Sym. & \std{94.50}{7.09} & \std{56.33}{4.31} & \underline{\std{55.67}{7.11}} & \std{57.17}{14.78} & \cellcolor{lightpink}\textbf{\std{50.33}{15.00}} & \std{65.67}{8.37} & \underline{\std{64.33}{3.75}} & \cellcolor{lightpink}\textbf{\std{62.83}{4.04}} & \cellcolor{lightpink}\textbf{\std{43.00}{2.78}} & \std{55.17}{4.19} & \underline{\std{44.83}{3.01}} \\
dishes & Vis. & \std{118.50}{14.08} & \std{118.17}{48.79} & \cellcolor{lightpink}\textbf{\std{105.67}{17.19}} & \std{133.83}{19.30} & \underline{\std{112.50}{12.12}} & \underline{\std{108.00}{8.00}} & \cellcolor{lightpink}\textbf{\std{96.17}{14.87}} & \std{118.83}{16.97} & \cellcolor{lightpink}\textbf{\std{71.83}{9.46}} & \std{105.67}{5.84} & \underline{\std{77.50}{17.76}} \\
\hline
Prepare & Sym. & \std{105.83}{1.26} & \std{67.33}{5.58} & \underline{\std{63.33}{10.26}} & \std{71.67}{1.53} & \cellcolor{lightpink}\textbf{\std{59.67}{0.29}} & \std{69.50}{5.89} & \underline{\std{63.83}{16.43}} & \cellcolor{lightpink}\textbf{\std{44.00}{4.00}} & \std{57.33}{9.57} & \underline{\std{50.33}{5.75}} & \cellcolor{lightpink}\textbf{\std{50.00}{6.50}} \\
meal & Vis. & \std{147.83}{16.78} & \std{99.17}{7.52} & \underline{\std{108.83}{10.79}} & \std{158.67}{5.01} & \cellcolor{lightpink}\textbf{\std{100.00}{18.03}} & \std{144.83}{26.42} & \underline{\std{92.83}{17.24}} & \cellcolor{lightpink}\textbf{\std{85.00}{3.54}} & \underline{\std{91.67}{8.31}} & \std{107.67}{14.15} & \cellcolor{lightpink}\textbf{\std{68.83}{5.77}} \\
\hline
Put & Sym. & \std{112.50}{0.50} & \std{72.33}{4.86} & \std{81.83}{5.03} & \underline{\std{72.33}{20.31}} & \cellcolor{lightpink}\textbf{\std{52.17}{1.53}} & \std{71.33}{7.91} & \underline{\std{57.50}{6.26}} & \cellcolor{lightpink}\textbf{\std{56.00}{4.44}} & \std{63.17}{7.01} & \cellcolor{lightpink}\textbf{\std{53.50}{2.18}} & \underline{\std{54.67}{1.89}} \\
groceries & Vis. & \std{158.17}{9.00} & \std{89.17}{10.20} & \underline{\std{118.83}{11.00}} & \std{146.00}{25.24} & \cellcolor{lightpink}\textbf{\std{76.17}{11.27}} & \std{144.50}{17.41} & \underline{\std{112.67}{6.53}} & \cellcolor{lightpink}\textbf{\std{67.33}{7.07}} & \std{105.67}{30.71} & \underline{\std{89.50}{12.26}} & \cellcolor{lightpink}\textbf{\std{83.83}{3.88}} \\
\hline
Set up & Sym. & \std{85.67}{6.93} & \std{54.50}{8.53} & \std{68.50}{4.09} & \underline{\std{64.33}{3.55}} & \cellcolor{lightpink}\textbf{\std{55.17}{5.75}} & \std{59.67}{4.04} & \cellcolor{lightpink}\textbf{\std{51.00}{1.00}} & \underline{\std{53.33}{0.58}} & \std{59.67}{3.62} & \underline{\std{55.50}{5.00}} & \cellcolor{lightpink}\textbf{\std{47.67}{1.04}} \\
table & Vis. & \std{109.17}{9.70} & \std{80.17}{9.46} & \underline{\std{98.50}{30.74}} & \std{130.17}{18.11} & \cellcolor{lightpink}\textbf{\std{93.00}{17.26}} & \std{110.83}{6.66} & \underline{\std{119.83}{15.33}} & \cellcolor{lightpink}\textbf{\std{84.00}{4.24}} & \std{90.67}{14.74} & \cellcolor{lightpink}\textbf{\std{71.50}{12.13}} & \underline{\std{72.17}{5.25}} \\
\midrule
\rowcolor{rcolor}
\multicolumn{2}{c|}{\textbf{Sym. Avg}} & \std{110.83}{1.97} & \std{68.90}{4.06} & \underline{\std{71.30}{2.70}} & \std{72.80}{0.61} & \cellcolor{lightpink}\textbf{\std{65.07}{1.88}} & \std{70.20}{1.39} & \underline{\std{59.07}{1.12}} & \cellcolor{lightpink}\textbf{\std{55.70}{0.78}} & \std{60.53}{2.25} & \underline{\std{59.90}{2.55}} & \cellcolor{lightpink}\textbf{\std{52.33}{1.56}} \\
\rowcolor{rcolor}
\multicolumn{2}{c|}{\textbf{Vis. Avg}} & \std{149.03}{5.88} & \std{101.63}{1.39} & \underline{\std{120.30}{3.34}} & \std{149.97}{7.11} & \cellcolor{lightpink}\textbf{\std{99.27}{5.23}} & \std{135.07}{7.35} & \underline{\std{107.13}{9.13}} & \cellcolor{lightpink}\textbf{\std{86.40}{1.41}} & \std{98.97}{3.91} & \underline{\std{98.67}{2.10}} & \cellcolor{lightpink}\textbf{\std{71.27}{1.26}} \\
\midrule

\multicolumn{13}{l}{\textit{Comm. Cost ($\downarrow$)}} \\
\midrule
Prepare & Sym. & --- & --- & \underline{\std{1137.83}{140.59}} & \std{5759.17}{839.20} & \cellcolor{lightpink}\textbf{\std{299.33}{20.53}} & \underline{\std{1248.67}{493.40}} & \std{6437.67}{847.38} & \cellcolor{lightpink}\textbf{\std{413.33}{249.15}} & \underline{\std{1089.67}{55.25}} & \std{13337.50}{2432.89} & \cellcolor{lightpink}\textbf{\std{303.50}{44.27}} \\
tea & Vis. & --- & --- & \underline{\std{1657.50}{371.05}} & \std{5293.67}{854.03} & \cellcolor{lightpink}\textbf{\std{306.83}{26.00}} & \underline{\std{1639.33}{638.09}} & \std{4842.00}{595.82} & \cellcolor{lightpink}\textbf{\std{301.67}{59.38}} & \underline{\std{972.83}{106.81}} & \std{8291.83}{1206.14} & \cellcolor{lightpink}\textbf{\std{308.00}{14.26}} \\
\hline
Wash & Sym. & --- & --- & \underline{\std{988.83}{103.83}} & \std{5649.83}{993.33} & \cellcolor{lightpink}\textbf{\std{277.50}{13.76}} & \underline{\std{1028.67}{354.98}} & \std{4304.83}{742.29} & \cellcolor{lightpink}\textbf{\std{241.67}{4.75}} & \underline{\std{683.00}{113.51}} & \std{11443.83}{2119.44} & \cellcolor{lightpink}\textbf{\std{336.50}{43.08}} \\
dishes & Vis. & --- & --- & \underline{\std{1096.67}{182.01}} & \std{3298.83}{1404.99} & \cellcolor{lightpink}\textbf{\std{304.67}{13.53}} & \underline{\std{1629.50}{439.98}} & \std{2954.00}{304.11} & \cellcolor{lightpink}\textbf{\std{224.00}{43.01}} & \underline{\std{600.67}{100.35}} & \std{6814.83}{552.12} & \cellcolor{lightpink}\textbf{\std{281.33}{6.79}} \\
\hline
Prepare & Sym. & --- & --- & \underline{\std{1667.33}{356.22}} & \std{9594.00}{1986.36} & \cellcolor{lightpink}\textbf{\std{292.00}{14.00}} & \underline{\std{1227.67}{246.36}} & \std{7686.50}{1268.34} & \cellcolor{lightpink}\textbf{\std{245.00}{8.19}} & \underline{\std{1116.67}{62.50}} & \std{15426.33}{1687.59} & \cellcolor{lightpink}\textbf{\std{305.83}{43.54}} \\
meal & Vis. & --- & --- & \underline{\std{1651.83}{114.86}} & \std{7015.00}{801.68} & \cellcolor{lightpink}\textbf{\std{262.00}{21.15}} & \underline{\std{1988.33}{692.49}} & \std{5807.33}{1464.37} & \cellcolor{lightpink}\textbf{\std{255.00}{11.72}} & \underline{\std{1222.17}{199.35}} & \std{10131.33}{4368.30} & \cellcolor{lightpink}\textbf{\std{269.67}{2.02}} \\
\hline
Put & Sym. & --- & --- & \underline{\std{1384.33}{216.43}} & \std{4791.00}{850.51} & \cellcolor{lightpink}\textbf{\std{322.83}{34.16}} & \underline{\std{922.00}{86.64}} & \std{4389.17}{456.48} & \cellcolor{lightpink}\textbf{\std{293.50}{31.19}} & \underline{\std{918.00}{172.44}} & \std{11486.67}{1507.05} & \cellcolor{lightpink}\textbf{\std{314.33}{52.78}} \\
groceries & Vis. & --- & --- & \underline{\std{1415.83}{196.08}} & \std{5574.33}{1574.59} & \cellcolor{lightpink}\textbf{\std{316.00}{50.47}} & \underline{\std{1671.33}{428.42}} & \std{4898.83}{356.46} & \cellcolor{lightpink}\textbf{\std{253.17}{21.01}} & \underline{\std{904.17}{115.67}} & \std{5815.17}{2296.64} & \cellcolor{lightpink}\textbf{\std{314.00}{25.10}} \\
\hline
Set up & Sym. & --- & --- & \underline{\std{1500.83}{61.31}} & \std{4931.50}{1097.43} & \cellcolor{lightpink}\textbf{\std{330.33}{70.67}} & \underline{\std{1126.50}{420.35}} & \std{3568.17}{467.03} & \cellcolor{lightpink}\textbf{\std{284.50}{44.72}} & \underline{\std{1157.50}{148.05}} & \std{4927.67}{1020.22} & \cellcolor{lightpink}\textbf{\std{307.83}{47.81}} \\
table & Vis. & --- & --- & \underline{\std{1396.33}{139.80}} & \std{2163.83}{768.13} & \cellcolor{lightpink}\textbf{\std{302.83}{32.20}} & \underline{\std{1086.67}{470.86}} & \std{2944.17}{607.16} & \cellcolor{lightpink}\textbf{\std{271.67}{27.19}} & \underline{\std{956.00}{258.02}} & \std{4440.67}{652.17} & \cellcolor{lightpink}\textbf{\std{277.83}{29.20}} \\
\midrule
\rowcolor{rcolor}
\multicolumn{2}{c|}{\textbf{Sym. Avg}} & --- & --- & \underline{\std{1335.83}{96.90}} & \std{6145.10}{330.51} & \cellcolor{lightpink}\textbf{\std{304.40}{18.74}} & \underline{\std{1110.70}{153.74}} & \std{5277.27}{323.72} & \cellcolor{lightpink}\textbf{\std{295.60}{58.58}} & \underline{\std{992.97}{11.82}} & \std{11324.40}{699.15} & \cellcolor{lightpink}\textbf{\std{313.60}{16.48}} \\
\rowcolor{rcolor}
\multicolumn{2}{c|}{\textbf{Vis. Avg}} & --- & --- & \underline{\std{1443.63}{84.60}} & \std{4669.13}{72.39} & \cellcolor{lightpink}\textbf{\std{298.47}{17.85}} & \underline{\std{1603.03}{248.26}} & \std{4289.27}{343.18} & \cellcolor{lightpink}\textbf{\std{261.10}{14.46}} & \underline{\std{931.17}{65.02}} & \std{7098.77}{1241.07} & \cellcolor{lightpink}\textbf{\std{290.17}{10.42}} \\
\bottomrule
\end{tabular}%
}
\label{tab:cross_model_method2_enlarged}
\end{table*}

\noindent\textbf{Baselines.} We select two types of baselines for performance comparison: traditional LLM-free agents and LLM-based agents. 
The traditional agents include: (i) MCTS-based Hierarchical Planner (MHP)~\citep{Zhang2023BuildingCE}:
% A hierarchical planning approach designed for the original Watch-And-Help Challenge. It features a Monte Carlo Tree Search (MCTS)-based high-level planner and a regression-based low-level planner. 
\duo{A hierarchical planning approach that features a Monte Carlo Tree Search (MCTS)-based high-level planner and a regression-based low-level planner.}
(ii) Rule-based Hierarchical Planner (RHP)~\citep{Zhang2023BuildingCE}: 
% A heuristic-based hierarchical planning approach designed for the original ThreeDWorld Transport Challenge. It uses a rule-based high-level planner combined with an A-start-based low-level planner for navigation.
\duo{A heuristic approach that uses a rule-based high-level planner combined with an A-start-based low-level planner for navigation.}
The LLM-based baselines include: (iii) CoELA~\citep{Zhang2023BuildingCE}: A collaboration framework based on step-by-step templated message generation and planning. (iv) CaPo~\citep{Liu2024CaPoCP}: A collaboration framework based on event-driven multi-round discussions.

% \noindent\textbf{Implementation details.} To evaluate CoBel-World across different underlying LLMs, we instantiate the LLM-based agents in CoBel-World and other LLM-based baselines using two state-of-the-art models: Qwen3-32B~\citep{yang2025qwen3}, an open-source model accessed via the Aliyun API, and ChatGPT-4o~\citep{hurst2024gpt}, a closed-source model accessed via the OpenAI API. We set the parameters with temperature = 0.7, top-p = 1, and a maximum token limit of 512 for both models. Unless otherwise stated, all experiments involve two agents on both benchmarks.

\noindent\duo{\textbf{LLM selection.} To comprehensively evaluate the effectiveness of CoBel-World across different LLMs, we adopt three state-of-the-art LLMs for CoBel-World and other LLM-based baselines: Qwen3-32B~\citep{yang2025qwen3}, DeepseekV3.2~\citep{liu2025deepseek} and GPT-4o~\citep{hurst2024gpt}. We set the temperature as 0.7, top-p as 1, and  maximum token limit as 512 for all LLMs. }

\subsection{Main Results}

\noindent\textbf{Performance.} 
% We summarize the performance of CoBel-World and baselines equipped with Qwen3-32B and ChatGPT-4o on the C-WAH and TDW-MAT benchmarks in Table \ref{tab:cross_model_method1} and Table \ref{tab:cross_model_method2}, respectively. 
Table \ref{tab:tdw_mat_updated_classic} and Table \ref{tab:cross_model_method2_enlarged} compare the performance of different methods on the TDW-MAT and C-WAH benchmarks, respectively.
%\zm{Notice that values highlighted in light pink denote the best performance of the task across all baseline methods; underlined values indicate the second-best results; rows shaded in light purple represent the average performance over a specific task category.}
% In general, LLM-based agents driven by 
%the small Qwen3-32B perform worse than traditional baselines due to the limited LLM model scale, but  agents powered by the more powerful 
% the powerful GPT-4o consistently outperform traditional baselines across most test settings, but agents driven by Qwen3-32B and DeepseekV3.2 exhibits variability compared to traditional baselines.
Our CoBel-World framework achieves superior task efficiency over all baseline methods while significantly reducing communication cost. On TDW-MAT, CoBel-World improves average transport rate by \textbf{\textit{4\%}} over the best baseline results; on C-WAH, it reduces average steps by \textbf{\textit{6-28\%}} compared to the strongest baseline. In terms of communication cost, CoBel-World reduces token usage by \textit{\textbf{64-79\%}} across all  settings. These results indicate that belief-driven collaboration not only reduces redundant communication but also enhances collaboration consistency. \zm{We also notice that LLM-based agents do not always outperform traditional agents. When driven by small LLMs like Qwen3-32B, CaPo is surpassed by RHP on TDW-MAT and both CaPo and CoELA are surpassed by MHP on C-WAH. In constrast, unlike these methods that highly rely on LLMs capabilities, CoBel-World  consistently achieves better performance, demonstrating its robustness.} % By comparison, baselines such as CoELA and CaPo rely on fixed communication protocols to exchange known information and thereby often fail to detect potential miscoordination until conflicting actions occur, leading to the drop of task completion efficiency. Moreover, they initiate communication even when collaboration is unnecessary (e.g., when agents can independently transport all objects in different rooms), causing higher communication cost.

\begin{figure*}[t]
    \centering
    \includegraphics[width=0.88\linewidth]{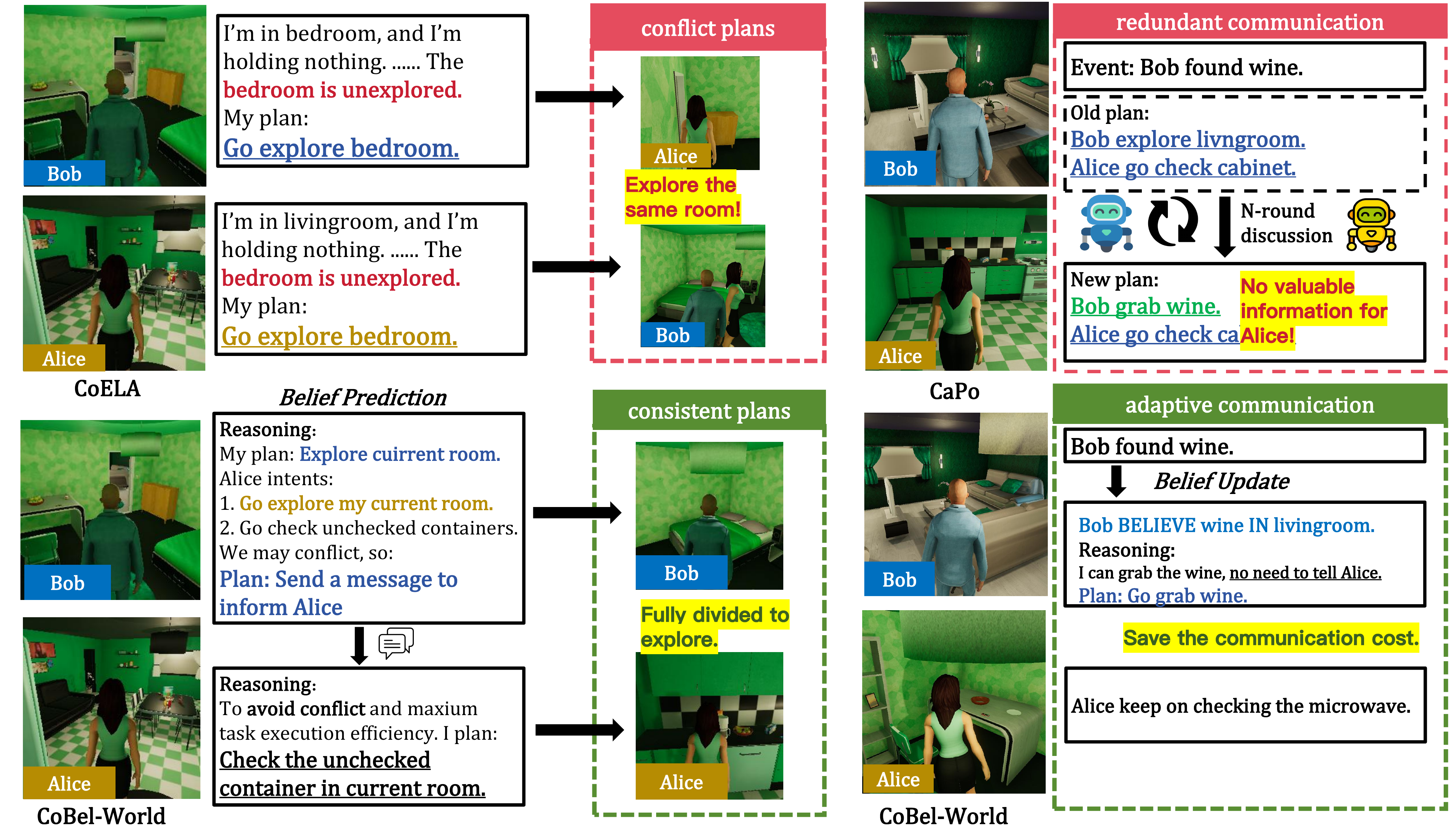} % 替换为你的图片路径
    \vspace{-0.05cm}
    % \caption{\zm{\textbf{Qualitative analysis of the advantages of CoBel-World over baseline methods.} To further analyze the advantages of CoBel-World, we compare its collaborative trajectories with those of CoELA and CaPo on the same task within the CWAH environment. Our evaluation focuses on two key dimensions: planning consistency and communication cost. The left part illustrates CoBel-World’s superior planning consistency over CoELA, while the right panel highlights its reduced communication compared to CaPo.}}
    \caption{\textbf{Illustration of the advantages of CoBel-World in terms of planning consistency and communication efficiency on C-WAH benchmark.} All methods are powered by GPT-4o. The left part illustrates CoBel-World’s superior planning consistency over CoELA, while the right panel highlights its reduced communication cost compared to CaPo.}
    \label{fig:qualitative}
\vspace{-0.2cm}
\end{figure*}

\noindent\textbf{Qualitative analysis.} 
% \zm{We sample collaboration trajectories of CoBel-World and two LLM-based baselines  CoELA and CaPo on the same task in the C-WAH environment, and analyze agents' collaboration behaviors to evaluate CoBel-World’s advantages in collaboration consistency and communication efficiency.}
Figure~\ref{fig:qualitative} illustrates the advantages of CoBel-World over baselines in terms of collaboration consistency and communication efficiency.
As shown in Figure~\ref{fig:qualitative} (left), at the initial stage of the task, agents will first make the plan. CoELA follows a fixed pipeline of communication-then-planning, which often fails to reach consensus with collaborators and leads to conflicting plans.
%(e.g. both agents explore the same room)
In contrast, CoBel-World performs belief prediction to infer the collaborators’ intents, detect potential miscoordination, and proactively initiate communication to reach consensus. For instance, Bob infers that Alice might explore his current room and thus proactively shares his intent and beliefs with her, enabling more consistent planning.
CaPo relies on event-triggered multi-round discussions to reach consensus with collaborators. However, when the triggering event provides little or no benefit to collaboration, this mechanism incurs unnecessary communication cost. As illustrated in Figure~\ref{fig:qualitative} (right), CaPo's discussions fail to yield better plans, resulting in redundant communication. In contrast, CoBel-World leverages belief modeling to autonomously assess the necessity of communication and dynamically decides whether to communicate to inform intents or directly execute a plan to maximize task efficiency.

\subsection{Ablation Study}
\textbf{Effects of each component.} 
% With C-WAH benchmark as example, we analyze the contributions of two key components in Cobel-World to collaboration: (1) \textit{Symbolic belief representation}, which enhances the LLMs' ability to represent and predict the joint states of environment and collaborators by converting partial observations into structured beliefs; (2) \textit{Bayesian belief prediction}, which infers collaborators’ intentions to perceive the collaborative status, thus guiding agents’ collaboration behaviors.  
% With C-WAH benchmark as example, we analyze the contributions of two key components in Cobel-World to collaboration: Symbolic Belief Representation (SBP) and Bayesian Belief Collaboration (BBC).
\duo{We analyze the contributions of two key components in Cobel-World to collaboration: symbolic belief representation (SBR) and Bayesian belief collaboration (BBC).} 
As shown in Table~\ref{Tab:ablation}, after removing the SBR module, Cobel-World exhibits a slight performance drop. This indicates that representing beliefs using unstructured natural language introduces redundant information, impairing LLMs’ planning capabilities. In contrast, emoving the BBC module leads to a severe performance drop. This phenomenon demonstrates that inferring collaborators’ intents  fosters more proactive collaboration.

% \noindent \textbf{Cobel-World with many agents.} Table~\ref{Tab:mang-agents} reports CoBel-World’s performance on the C-WAH benchmark as the number of agents scales beyond two. A significant performance gain is observed when scaling from two to three agents. However, increasing the agent number to four yields only marginal improvements in average steps.  This is because the C-WAH benchmark includes a number of relatively simple tasks composed of only 2–3 subgoals and thus cannot fully leverage the capacity of four agents. As the “wash dishes” task illustrated in the Appendix \ref{cwah}, only two objects require collection and transport, making collaboration among more than two agents unnecessary and potentially hinder consistent planning.

\noindent \duo{\textbf{Scaling to more agents.} To validate the scalability of CoBel-World in scenarios with larger teams, we report its performance as the number of agents increases. As shown in Table~\ref{Tab:mang-agents}, CoBel-World achieves significant performance gains when adding more agents, showing strong scalability.}
% \vspace{-0.2cm}

\noindent \textbf{Cobel-World with heterogeneous agents.}
% As shown in Table \ref{Tab:heterogeneous-exp}, pairing a CoBel-World agent with a CoELA agent significantly outperforms the homogeneous CoELA-CoELA setting, although it still falls short of the pure CoBel-World setting. This suggests that the belief modeling capability integrated into CoBel-World enables agents to proactively coordinate with heterogeneous collaborators, facilitating human-like collaboration.
\duo{As shown in Table~\ref{Tab:heterogeneous-exp}, a team comprising one CoBel-World agent and one CoELA agent substantially outperforms the homogeneous CoELA-CoELA setting, though it does not fully match the performance of the pure CoBel-World team. This demonstrates that CoBel-World’s structured belief modeling enables effective coordination with heterogeneous partners, thereby supporting flexible collaborative scenarios.}
\begin{table}[t] % [!ht] 强制尝试放在当前位置，放不下再考虑页面顶部
    \centering
    \small
    \caption{Effects of the components in CoBel-World using GPT-4o on C-WAH benchmark. Average steps required to complete task are reported. ``SBR'' denotes ``symbolic belief representation'' and ``BBC'' denotes ``Bayesian belief collaboration''.}
    \label{Tab:ablation}
    \vspace{-0.25cm}
    \begin{tabular}{lc}
        \toprule
        Method & Symbolic Obs ($\downarrow$) \\
        \midrule
        CoBel-World           & \textbf{52} \\
        CoBel-World (No SBR)  & 55 \\
        CoBel-World (No BBC)  & 68 \\
        \bottomrule
    \end{tabular}
\end{table}

% 表格 2：智能体数量实验
\begin{table}[!ht]
    \centering
    \small
    \caption{Benefits of increasing agent number in our CoBel-World using GPT-4o on {C-WAH benchmark}. Average steps required for task completion are reported.}
    \label{Tab:mang-agents}
    \vspace{-0.25cm} % 稍微调整间距，避免与上方表格贴得太近
    \begin{tabular}{l m{2.5cm}<{\centering} }
        \toprule
        Method & Symbolic Obs ($\downarrow$) \\
        \midrule
        CoBel-World$\times$2 & 52 \\
        CoBel-World$\times$3 & 47 \\
        CoBel-World$\times$4 & \textbf{43}\\
        \bottomrule
    \end{tabular}
\end{table}

\begin{table}[!ht]
    \centering
    \small
    \caption{Performance comparison of heterogeneous agent collaboration on the TDW-MAT benchmark.}
    \label{Tab:heterogeneous-exp}
    \vspace{-0.25cm}
    \begin{tabular}{l c c c}
        \toprule
        Method & Food & Stuff & Total \\
        \midrule
        CoELA + CoELA & 63 & 63 & 63 \\
        CoELA + CoBel-World & 75 & 71 & 73 \\
        CoBel-World + CoBel-World & 81   & 79 & \textbf{80} \\
        \bottomrule
    \end{tabular}
\end{table}
%%%%%%%%%%%%%%%%%%%%%%%%%%%%%%%%%%%%%%%

%%%%%%%%%%%%%%%%%%%%%%%%%%%%%%%%%%%%%%%
\section{Concluding Remarks}
\label{concluding_remarks}
% In this work, we introduced CoBel-World, a framework that equips LLM-based embodied agents with a \textit{collaborative belief world} to enable efficient and consistent multi-agent collaboration under partial observability. CoBel-World formalizes world and mental state knowledge into a structured symbolic belief language and leverages LLMs' zero-shot reasoning for Bayesian-style belief updates. With CoBel-World, LLM agents can proactively infer teammates' intentions and detect potential miscoordination. This intent-aware belief modeling supports adaptive communication, generating messages only when necessary to resolve conflicts or align critical information, thereby reducing redundant dialogue and physical actions.
% Extensive experiments on challenging benchmarks (TDW-MAT and C-WAH) show that CoBel-World reduces communication costs by 22–60\% while consistently improving task completion efficiency over state-of-the-art baselines. These results validate that explicit belief representation is a key enabler of scalable and human-like collaboration in open-ended environments.

\duo{In this work, we introduce CoBel-World, a framework that equips LLM-based agents with a collaborative belief world to enable efficient and consistent multi-agent collaboration under partial observability. CoBel-World first uses a symbolic belief representation module to translate linguistic descriptions of open-ended world into structured beliefs, then harnesses LLM reasoning to perform Bayesian-style belief updates in a zero-shot manner. With CoBel-World, LLM agents can  proactively infer teammates' intentions, adaptively communicate with others  and detect potential miscoordination, thereby reducing redundant dialogue and physical actions.
Extensive experiments show that CoBel-World reduces communication cost by 64-79\% while consistently improving task completion efficiency over state-of-the-art baselines. }
%%%%%%%%%%%%%%%%%%%%%%%%%%%%%%%%%%%%%%%

%%%%%%%%%%%%%%%%%%%%%%%%%%%%%%%%%%%%%%%
\section{Limitations}
\label{limitations}
% \noindent \textbf{Sensitivity to the reasoning stability of LLMs.} 
\noindent \duo{\textbf{Sensitivity to hallucinations.} }
% The performance of CoBel-World is highly relavant to the reasoning ability of the underlying LLM. Since our framework uses LLM reasoning to perform zero-shot belief updates, minor cognitive fluctuations or logical inconsistencies during the reasoning process may undermine its overall stability. Our analysis of failure cases (in Appendix \ref{appendix:failure_cases}) shows that LLM hallucinations may lead to incorrect symbolic beliefs, which then influence the subsequent task execution. Future research may incorporate self-reflection or automated verification mechanisms to bolster the robustness of LLM-driven reasoning.
% The performance of CoBel-World is highly relevant to the reasoning ability of the underlying LLM. 
Since our framework uses LLM reasoning to perform zero-shot belief updates, hallucinations during the reasoning process may degrade performance. As shown in our failure analysis (Appendix \ref{appendix:failure_cases}), hallucinations can lead to incorrect symbolic beliefs, which then influence the subsequent task execution. Future research may incorporate self-reflection or automated verification mechanisms to enhance the robustness of belief generation.

\noindent \textbf{Lack of multimodal reasoning.} 
% While the benchmarks provide high-dimensional visual observations, CoBel-World follows the settings in prior works—transforming the visual inputs into textual descriptions to serve as the agent's observations. Such transformation limits the agent's ability to reason directly over raw visual features to facilitate fine-grained collaboration. Future work could explore the deployment of Multimodal Large Language Models (MLLMs) to drive the agents, thereby enabling direct visual belief reasoning and potentially achieving a more granular understanding of complex, open-world environments.
Although the evaluation  benchmarks provide high-dimensional visual observations, CoBel-World adopts the common practice in prior work--converting raw visual inputs into textual scene descriptions to serve as agent observations. While this abstraction simplifies reasoning, it fails to directly exploit the fine-grained visual cues. Future work could leverage multimodal LLMs to enable agents to perform belief updates directly from visual inputs to facilitate more effective collaborative behaviors in complex, open-world environments.
%%%%%%%%%%%%%%%%%%%%%%%%%%%%%%%%%%%%%%%

%%%%%%%%%%%%%%%%%%%%%%%%%%%%%%%%%%%%%%%
\section{Ethical Considerations}
\label{ethical_considerations}
In developing CoBel-World, we have considered several ethical points regarding the deployment of LLM-based agents:

\noindent \textbf{Potential for model bias}: Since our framework relies on LLMs for intent inference, the agents may inherit social or behavioral biases from the models' pre-training data. We encourage developers to monitor these behaviors in human-robot interaction scenarios.

\noindent \textbf{Environmental impact:} The use of large-scale LLMs for continuous reasoning and planning requires significant computational power. We suggest that future research explore the use of smaller, task-specific models to reduce the energy consumption and carbon footprint of these systems.
%%%%%%%%%%%%%%%%%%%%%%%%%%%%%%%%%%%%%%%

%%%%%%%%%%%%%%%%%%%%%%%%%%%%%%%%%%%%%%%
\section{Author Contribution}
\textbf{Zhimin Wang} initiated and led the project. He is responsible for for the entire process from research motivation, methodology design, to experiments design. He contributed to the implementation of CoBel-World's core component. In paper writing, he completed most of the writing and revision work of the paper.

\noindent \textbf{Duo Wu} provided extensive supervision throughout the project, including guidance on problem definition, idea refinement, experiments design and paper writing. He also contributed to the revision of the paper.

\noindent \textbf{Shaokang He} reproduced the baseline methods and the partial implementation of CoBel-World. He also contributed to performance optimization and conducted large-scale experiments.

\noindent \textbf{Jinghe Wang} contributed technical solutions for system performance optimization.

\noindent \textbf{Linjia Kang}, \textbf{Jing Yu}, \textbf{Kai Zhu} and \textbf{Jiawei Li} assisted with manuscript proofreading and language polishing.

\noindent \textbf{Zhi Wang} provided overall supervision, institutional resources, and financial support for the project.

\bibliography{latex/cobel-world}

%%%%%%%%%%%%%%%%%%%%%%%%%%%%%%%%%%%%%%%
\appendix
\label{sec:appendix}
\newpage

\section{Code Availability and Reproducibility}
\label{appendix:code}
To facilitate the reproducibility of our results, we provide the full source code and implementation details of \textbf{CoBel-World} in the following anonymous repository: \\
\url{https://anonymous.4open.science/r/CoBel_World}

\section{Theoretical Analysis of CoBel-World}
\label{appendix:formulation}
\subsection{Multi-Agent Collaboration Formulation}
\label{appendix:decmdp}
We model the multi-agent collaboration task as a \textit{decentralized partially observable Markov decision process (DEC-POMDP)} \citep{Oliehoek2016ACI, Bernstein2000TheCO, Spaan2006DecentralizedPU}, defined by the tuple:\[
\mathcal{M} = \langle I, \mathcal{S}, \{\mathcal{A}_i\}, \{\mathcal{O}_i\}, T, \{O_i\}, R, h \rangle,
\]  
where:  
\begin{itemize}
    
\item $ \mathcal{N} = \{ n_0, n_1, ... n_m \}  $ is a finite set of $ m $ agents;
    \item $ \mathcal{S} $ is a finite state space representing the environment;
    \item $ \mathcal{A}_i $ is the action set of agent $ n_i $, with $ \mathcal{A} = \times_{i \in I} \mathcal{A}_i $ the joint action space;
    \item $ \mathcal{O}_i $ is the observation set of agent $ n_i $, encompassing partial ego-centric visual inputs and received messages;
    \item $ T(s' \mid s, \mathbf{a}) = p(s' \mid s, \mathbf{a}) $ is the transition function, denoting the probability of transitioning to state $ s' \in \mathcal{S} $ from $ s \in \mathcal{S} $ under joint action $ \mathbf{a} \in \mathcal{A} $;
    \item $ O_i(o_i \mid s', \mathbf{a}) = p(o_i \mid s', \mathbf{a}) $ is the observation model for agent $ i $, giving the probability of observing $ o_i \in \mathcal{O}_i $ upon reaching $ s' $ after executing $ \mathbf{a} $;
    \item $ R(s, \mathbf{a}) $ is the global reward function shared by all agents;
    \item $ h $ is the finite planning horizon. 
\end{itemize}
The objective is for the team to maximize the expected cumulative reward $ \mathbb{E} \left[ \sum_{t=0}^{h-1} R(s^t, \mathbf{a}^t) \right] $ through decentralized execution of a joint policy $ \pi = \{\pi_i\}_{i \in I} $, where each agent $ i $ selects actions $ a_i^t \sim \pi_i(\cdot \mid \tau_i^t) $ based only on its local observation-action history $ \tau_i^t = (o_i^0, a_i^0, \dots, o_i^t) $.

\subsection{Belief Update with Bayesian filter.}
\label{belieffilter}
Due to partial observability, each agent $ n_i $ maintains a \textit{belief state} $ b_i: \mathcal{S} \to [0,1] $, which represents its subjective probability distribution over the true state $ s \in \mathcal{S} $. The belief $ b_i^t $ at time $ t $ is conditioned on the agent's local history $ \tau_i^t = (o_i^0, a_i^0, \dots, o_i^t) $. Upon executing action $ a^t \in \mathcal{A} $ and receiving observation $ o_i^{t+1} \in \mathcal{O}_i $, agent $ i $ updates its belief using a Bayesian filter:

\begin{equation}
    b_i'(s') \propto O_i(o_i' \mid s', \mathbf{a}) \sum_{s \in \mathcal{S}} T(s' \mid s, \mathbf{a}) \, b_i(s),
\end{equation}
where $ b_i $ is the current belief, $ b_i' $ is the updated belief, $ \mathbf{a} $ is the joint action, $ o_i' $ is the new observation, and $ T $ and $ O_i $ are the transition and observation models, respectively.
% This process decomposes into a \textit{prediction step} and a \textit{measurement update}:
% \begin{align}
%     \text{(Prediction)} \quad & b_i^{\text{pred}}(s') = \sum_{s \in \mathcal{S}} T(s' \mid s, \mathbf{a}) \, b_i(s), \\
%     \text{(Measurement update)} \quad & b_i'(s') \propto O_i(o_i' \mid s', \mathbf{a}) \, b_i^{\text{pred}}(s'),
% \end{align}
% where $b_i$ denotes agent $i$'s belief over the environment state. In CoBel-World, we instantiate this framework by leveraging LLMs to perform the prediction step (reasoning about collaborators' intents) and use direct observations (vision and communication) for the measurement update. This Bayesian-style reasoning forms the core of our collaborative belief modeling.
This update decomposes into two conceptually distinct stages:

\noindent\textbf{Prediction.} The agent predict possible beliefs based on its current belief:
\begin{equation}
    \overline{b}_i(s') = \sum_{s \in \mathcal{S}} T(s' \mid s, \mathbf{a}) \, b_i(s),
\end{equation}
resulting in a prior belief $ \overline{b}_i $ that captures the expected state distribution after the action. In our framework, this step is enhanced by \textit{theory of mind reasoning}~\citep{Li2023TheoryOM,ma2023towards}, enabling agents to anticipate teammates' intentions.

\noindent\textbf{Measurement update.} The agent conditions the prior on the new observation $ o_i' $ (including visual input and messages):
\begin{equation}
    b_i'(s') \propto O_i(o_i' \mid s', \mathbf{a}) \cdot \overline{b}_i(s'),
\end{equation}
yielding a posterior belief $ b_i' $ that incorporates direct evidence. This step enables rapid belief alignment through perception and communication.

This Bayesian-style process--predicting future states and update based on observations--forms the theoretical foundation of {{CoBel-World}}. 

\section{Additional Environment Details}
\label{env}
\duo{Following CoELA~\citep{Zhang2023BuildingCE}}, we evaluate our methods and baselines on two challenging embodied multi-agent benchmarks with open-ended environments: ThreeDWorld Multi-Agent Transport (TDW-MAT) and Communicative Watch-And-Help (C-WAH). Detailed descriptions of these benchmarks are provided below. 

\subsection{TDW-MAT}
\label{tdw}

% \noindent\textbf{Tasks.} TDW-MAT comprises two distinct task categories: food-transportation and object-transportation. The food-transportation task involves 6 types of  target objects including apple, banana, orange, bread, loaf bread, burger; and three container types: bowl, plate, and tea tray. And the object-transportation task includes another 6 different target objects including calculator, mouse, pen, lighter, purse, iPhone; and  three container types: plastic basket, wooden basket, and wicker basket. In each task instance, the environment contains 10 target objects and between 2 to 5 containers. The scenes are structured across four semantically coherent room types: living room, office, kitchen, and bedroom with object placements adhering to real-world contextual plausibility. Agents are required to maximize the number of target objects delivered to a designated goal location within a time budget of 3,000 simulation frames. Containers serve as transport tools, each capable of carrying up to three objects; in their absence, agents can carry at most two objects simultaneously.

\duo{\noindent\textbf{Tasks.} The test set of TDW-MAT consists of 24 episodes, evenly divided into two task categories: food transportation and object transportation.
The food transportation tasks involves:}
\begin{itemize}
    \item 6 types of target objects: apple, banana, orange, bread, loaf, and burger.
    \item 3 types of containers: bowl, plate, and tea tray.
\end{itemize}
The object transportation tasks includes:
\begin{itemize}
    \item 6 types of target objects: calculator, mouse, pen, lighter, purse and iPhone.
    \item 3 types of containers: plastic, wooden, and wicker baskets.
\end{itemize}
\duo{In each task instance, the environment contains 10 target objects and 2-5 containers.  Scenes are instantiated across four semantically coherent room types: living room, office, kitchen, and bedroom, with object placements adhering to real-world contextual plausibility.
Agents are required to maximize the number of target objects delivered to a designated goal location within a time budget of 3,000 simulation frames. Containers serve as transport tools, each capable of carrying up to three objects; Without a container, each agent can carry at most two objects simultaneously.}

\noindent\textbf{Observation space.} The embodied agent receives a variety of observations, with the primary ones being an egocentric RGB image and a depth image. Additionally, there are several auxiliary observations. The observation space includes:
\begin{itemize}
    \item {\textbf{RGB image.} An ego-centric image captured by a forward-facing camera, with a resolution of 512 × 512 and a field of view of 90 degrees.}
    \item {\textbf{Depth image.}} This image shares the same camera intrinsic parameters as RGB image.
    \item {\textbf{Oracle perception (optional). }} An image where each object ID is represented by a distinct color, using the same camera intrinsic parameters as the RGB image. 
    \item {\textbf{Agent position and rotation. }} The position and rotation of the agent within the simulation environment.
    \item {\textbf{Messages. }} Information sent by all agents.
    \item {\textbf{Held objects. }} Information about the objects currently held by the agent.
    \item {\textbf{Opponent held objects. }}  Information about objects held by another agent, if the agent is within view.
\end{itemize}
\noindent\textbf{Action space.} In TDW-MAT, agents can perform 7 distinct types of actions to interact with the environment or communicate with each other. Each action spans multiple frames. The detailed action space is outlined below:
\begin{itemize}
    \item {\textbf{Move forward. }} The agent advances by 0.5m.
    \item {\textbf{Turn left. }} The agent rotates left by 15 degrees.
    \item {\textbf{Turn right. }}  The agent rotates right by 15 degrees.
    \item {\textbf{Grasp. }} The agent grasps an object, successfully performing this action only when in close proximity to the object. The object can be either a target or a container.
    \item {\textbf{Put in. }}The agent places a target into a container, an action that is possible only when the agent is holding a target in one hand and a container in the other.
    \item  {\textbf{Drop. }} The agent releases the objects held in hand.
    \item {\textbf{Send message. }} The agent sends a message to others, with a limit of 500 characters per frame.
\end{itemize}

% \begin{table*}[!h]
%     \centering
%     \caption{TDW_MAT tasks extended with capacity dimension}
%     \vspace{-0.1cm}
%     \label{tab:capacity}
%     \resizebox{\textwidth}{!}{%
%     \begin{tabular}{C{4cm}cll}
%     \toprule
%     \multicolumn{1}{c}{Task Type} &  Container Num & \multicolumn{1}{c}{Container Name}\\
%     \midrule
%     Food-low-capacity & 2 & tea tray, bowl, plate \\
%     \midrule
%     Food-high-capacity & 5 & tea tray, bowl, plate \\
%     \midrule
%     Stuff-low-capacity & 2 & wood basket, wicker basket, plastic basket \\
%     \midrule
%     Stuff-high-capacity & 5 & wood basket, wicker basket, plastic basket  \\
%     \bottomrule
%     \end{tabular}}
%     \vspace{-0.1cm}
% \end{table*}

\begin{table*}[t] % 跨栏建议用 [t]
    \centering
    \caption{TDW_MAT tasks extended with capacity dimension}
    \label{tab:capacity}
    % 使用 tabularx，总宽设为 \textwidth
    \small
    \begin{tabularx}{0.8\textwidth}{l c X} % X 列会自动伸展填充剩余空间
    \toprule
    Task Type & Container Num & Container Name \\
    \midrule
    Food-low-capacity & 2 & tea tray, bowl, plate \\
    \midrule
    Food-high-capacity & 5 & tea tray, bowl, plate \\
    \midrule
    Stuff-low-capacity & 2 & wood basket, wicker basket, plastic basket \\
    \midrule
    Stuff-high-capacity & 5 & wood basket, wicker basket, plastic basket \\
    \bottomrule
    \end{tabularx}
\end{table*}

\noindent\textbf{Extended TDW-MAT tasks.} Building upon the classic TDW-MAT benchmark introduced by CoELA~\citep{Zhang2023BuildingCE}, we extend the evaluation along task difficulty dimension to enable a more comprehensive comparison between CoBel-World and various baselines. Specifically, tasks are categorized into low-capacity and high-capacity settings based on the number of containers available to the agent in the environment. Each difficulty level comprises half of both the food-transportation and stuff-transportation tasks. Task details are provided in Table~\ref{tab:capacity}.

\subsection{C-WAH}
\label{cwah}
C-WAH builds upon the Watch-And-Help challenge~\citep{puig2020watch} by incorporating the ability for agents to send messages to one another. Sending messages, like other actions, consumes one timestep and is subject to a maximum length constraint.

\begin{table*}[t]
    \centering
    \caption{Detailed description of C-WAH tasks}
    \label{tab:cwah_des}
    \vspace{-0.1cm}
    \small
    \begin{tabular}{R{3.0cm}cc}
    \toprule
     \multicolumn{1}{c}{Task Name}  &  \multicolumn{2}{c}{Oject Set}\\
     \midrule
    Prepare afternoon tea   &  \multicolumn{2}{l}{ON(cupcake,coffeetable), ON(pudding,coffeetable),}\\
    ~ &  \multicolumn{2}{l}{ON(apple,coffeetable), ON(juice,coffeetable),}\\
    ~ &  \multicolumn{2}{l}{ON(wine,coffeetable)}\\
    \midrule
    Wash dishes & \multicolumn{2}{l}{IN(plate,dishwasher), IN(fork,dishwasher)}\\
    \midrule
    Prepare a meal & \multicolumn{2}{l}{ON(coffeepot,dinnertable),ON(cupcake,dinnertable),}\\
    ~ & \multicolumn{2}{l}{ON(pancake,dinnertable), ON(poundcake,dinnertable),}\\
    ~ & \multicolumn{2}{l}{ON(pudding,dinnertable), ON(apple,dinnertable),}\\
    ~ & \multicolumn{2}{l}{ON(juice,dinnertable), ON(wine,dinnertable)}\\
    \midrule
    Put groceries & \multicolumn{2}{l}{IN(cupcake,fridge), IN(pancake,fridge),}\\
    ~ & \multicolumn{2}{l}{IN(poundcake,fridge), IN(pudding,fridge),}\\
    ~ & \multicolumn{2}{l}{IN(apple,fridge), IN(juice,fridge),}\\
    ~ & \multicolumn{2}{l}{IN(wine,fridge)}\\
    \midrule
    Set up a dinner table & \multicolumn{2}{l}{ON(plate,dinnertable), ON(fork,dinnertable)}\\
    \bottomrule
    \end{tabular}
    \vspace{-0.2cm}
\end{table*}

% \begin{table*}[t]
%     \centering
%     \caption{Detailed description of C-WAH tasks}
%     \label{tab:cwah_des}
%     % 使用 tabularx，将 Object Set 这一列设为 X
%     \begin{tabularx}{0.9\textwidth}{R{3.0cm} X} 
%     \toprule
%     Task Name & Object Set \\ % 修正了 Object 拼写
%     \midrule
%     Prepare afternoon tea & ON(cupcake,coffeetable), ON(pudding,coffeetable), ON(apple,coffeetable), ON(juice,coffeetable), ON(wine,coffeetable) \\
%     \midrule
%     Wash dishes & IN(plate,dishwasher), IN(fork,dishwasher) \\
%     \midrule
%     Prepare a meal & ON(coffeepot,dinnertable), ON(cupcake,dinnertable), ON(pancake,dinnertable), ON(poundcake,dinnertable), ON(pudding,dinnertable), ON(apple,dinnertable), ON(juice,dinnertable), ON(wine,dinnertable) \\
%     \midrule
%     Put groceries & IN(cupcake,fridge), IN(pancake,fridge), IN(poundcake,fridge), IN(pudding,fridge), IN(apple,fridge), IN(juice,fridge), IN(wine,fridge) \\
%     \midrule
%     Set up a dinner table & ON(plate,dinnertable), ON(fork,dinnertable) \\
%     \bottomrule
%     \end{tabularx}
% \end{table*}

\noindent{\textbf{Tasks.}} C-WAH  comprises five household-oriented tasks: preparing afternoon tea, Washing dishes, preparing a meal, putting groceries, and setting up a dinner table. The test set contains 10 episodes, including both symbolic and visual observation settings. Each task involves multiple subtasks, expressed through predicates in the form “ON/IN(x, y)”, which correspond to actions like “Place x ON/IN y”. Detailed information is provided in Table \ref{tab:cwah_des}. 
The primary objective is to complete all given subtasks within 250 timesteps, with each task containing 3-5 subtasks.

\noindent\textbf{Observation space.}  C-WAH  provides two observation modalities: symbolic observation and visual observation.
In symbolic observation, agents have full access to all object-related information in the same room, including object  names, locations, states, and relational attributes.
In visual observation, agents receive ego-centric RGB and depth images along with auxiliary observations.  Detailed observations include:
\begin{itemize}
    \item {\textbf{RGB image. }}An ego-centric image from a forward-facing camera, with a resolution of 256 × 512 and a field of view of 60 degrees.
    \item {\textbf{Depth image. }} An image with the same camera intrinsic parameters as the RGB image.
    \item {\textbf{Oracle Perception. }} An image where each object ID is mapped to a color, sharing the same camera intrinsic parameters as the RGB image.
    \item {\textbf{Agent position. }} The agent’s position within the simulation world.
    \item {\textbf{Messages. }} Information sent by all agents.
    \item {\textbf{Held objects. }}Information about the objects currently held by the agent.
    \item {\textbf{Opponent held objects. }}Information about objects held by another agent, if visible.
    
\end{itemize}
\noindent{\textbf{Action space.}} The action space in C-WAH includes:
\begin{itemize}
    \item {\textbf{Walk towards. }}Move towards an object in the same room or towards a specific room.
    \item {\textbf{Turn left. }} Rotate left by 30 degrees.
    \item {\textbf{Turn right.  }} Rotate right by 30 degrees.
    \item {\textbf{Grasp. }}  Grasp an object, which can be successfully performed only when the agent is close to the object.
    \item{\textbf{Open. }}Open a closed container, performable only when the agent is near the container.
    \item {\textbf{Close. }}Close an open container, performable only when the agent is near the container.
    \item{\textbf{Put.  }} Place held objects into an open container or onto a surface, performable only when the agent is near the target position.
    \item {\textbf{Send message. }}  Communicate with others, with a limit of 500 characters per message.
\end{itemize}

\section{Cobel-World Details}
\label{appendix-cobel}
\subsection{Belief Symbolic Representation}
\label{beliefrepresentation}

\noindent\textbf{Prompt templates.} 
% We list the prompts used in the Collaborative Representing Protocol. In this protocol, two agents iteratively perform a propose-and-revise procedure based on the task description until a reviewer verifies that the resulting representation is logically consistent and complete in terms of agent roles and key entities.
We list the belief rules construction prompts for the two agents Alice and Bob in the benchmarks, as shown in Figure~\ref{fig:alice_bcp} and Figure~\ref{fig:bob_bcp}, respectively.

\noindent\textbf{Belief rules.} Figure~\ref{fig:belief_rules} illustrates the belief rules of CoBel-World.
% \afterpage{
\begin{figure*}[t]
\centering
\begin{tcolorbox}[ colback=Emerald!10,
    colframe=cyan!40!black,
    title=\textbf{Belief Rules Construction Prompt of Alice}]
    \textbf{Init Prompt:} You are Alice, you and Bob are constructing beliefs rules to denote the zero and first order belief of the world. You should first extract entity types and predicates in a specific domain given a task description and the belief symbolic language below. After that you should use the belief symbolic language to describe the possible belief types in this task domain and send to bob for discussion.\\

Belief symbolic language:~ \$BELIEF_LANGUAGE \$\\

Task description:~\$TASK_DESCRIPTION\$\\
Note that the zeroth-order belief denote my knowledge of the world, first-order belief denote my knowledge of others belief.\\
DO NOT generate beliefs that go beyond the information specified in the task description.
Consider ONLY zero-order and first-order beliefs.

The belief rules should be in syntax format with entity represented with a "?" prefix, and without any additional comment and analysis and explanation:
You should output strictly in the format of the following structure:\\
\\
{Entity and predicate reasoning:}\\
{Zero order belief rules:}\\
{First order belief rules:}\\

\textbf{Refine Prompt: }You are Alice, you and Bob are constructing beliefs rules to denote the zero and first order belief of the world. Given a task description and the belief symbolic language below, you should refine the belief rules according to Bob's suggestions.\\

Belief symbolic language:~\$BELIEF_LANGUAGE\$\\
Task description:~\$TASK_DESCRIPTION\$\\
previous content:~\$PREVIOUS_CONTENT\$\\
Bob's suggestions:~\$SUGGESTIONS\$\\

DO NOT generate beliefs that go beyond the information specified in the task description.\\
Consider ONLY zeroth-order and first-order beliefs.\\
Note that the zeroth-order belief denote my knowledge of the world, first-order belief denote my knowledge of others belief.\\
Now try to refine your previous output according to Bob's suggestions.
The belief rules should be in syntax formatwith entity represented with a ? prefix, and without any additional comment and analysis and explanation:
You should output strictly in the format of the following structure:\\\\
Reasoning: \\
Zero order belief rules:\\
First order belief rules:\\
\end{tcolorbox}
    \caption{Alice's belief rules construction prompt}
    \label{fig:alice_bcp}
\end{figure*}
% }

\subsection{Bayesian Belief Collaboration}
\label{bayesianbeliefcollaboration}
In this part, we list the prompts used in the Bayesian belief collaboration module on TDW-MAT benchmark. Figure~\ref{fig:belief-update-zero} and Figure~\ref{fig:zero-order-predict} illustrate the prompts for zero-order belief update and prediction, respectively. Figure~\ref{fig:first-belief-update} and Figure~\ref{fig:first-belief-prediction} illustrate the prompts for first-order belief update and prediction, respectively. Figure~\ref{fig:adaptivecollaboration}, Figure~\ref{fig:communication}, Figure~\ref{fig:planning} and Figure~\ref{fig:replanning} depict the prompts for adaptive collaboration, communication, planning and replanning, respectively.

\subsection{Failure cases}
\label{appendix:failure_cases}
The failures from CoBel-World primarily stem from inherent hallucinations in LLMs. Although our symbolic belief language substantially reduces such hallucinations, it cannot fully eliminate them. We give detailed failure cases in Figure~\ref{fig:failure_case}.

\begin{figure*}[h]
\centering
\begin{tcolorbox}[ colback=Emerald!10,
    colframe=cyan!40!black,
    title=\textbf{Failure Cases}]
\textbf{\textit{Failure type:}} \textcolor{blue}{belief update errors}

\textbf{\textit{Failure log:}}

\texttt{Alice: low-level-plan: send a message: Hi Bob, I've explored part of the <Office>(3000). My subplan is to continue exploring the <Office>(3000) and transport objects to the bed.}

\texttt{belief_episode_0 - INFO - first belief update:}

\texttt{first-order beliefs: }

\texttt{\textcolor{red}{Bob BELIEVE Alice BELIEVE <bed> IN <Office> (3000)}}

\textbf{\textit{Analysis:}} In this case, LLM mistakenly associated \texttt{bed} and \texttt{office (3000)} in a sentence as belief, and finally got the wrong belief.

\

\textbf{\textit{Failure type:}} \textcolor{blue}{belief prediction errors}

\textbf{\textit{Failure log:}}

\texttt{Bob predict_first: \textcolor{red}{The <Kitchen> (5000) is commonly associated with food items} like oranges, breads, and bananas. Exploring this room might yield a higher chance of finding the goal objects.}

\texttt{Bob oppo_subplans: }

\texttt{- go to <Kitchen>(5000)}
\texttt{- explore current room <Kitchen>(5000)}

\textbf{\textit{Analysis:}} LLM mistakenly used common sense as a priori knowledge to aid belief prediction, but there is no setting in the experimental setting that "food is more likely to be in the kitchen", as a result, LLM suggest go to kitchen first. 

\

\textbf{\textit{Failure type:}} \textcolor{blue}{action-selection errors}

\textbf{\textit{Failure log:}}

\texttt{Alice intuitive_planning: The last time you saw **Bob was in the <Office> (1000)**, which hasn't been explored yet. It may contain target objects or provide useful information for the goal.}

\texttt{\textcolor{red}{Answer: A. go to <Office> (1000)}}

\textbf{\textit{Analysis:}} Although LLM believed that Bob is in the \texttt{<office>(1000)} , LLM still chose to go to the \texttt{<office>(1000)} which is a redundant action.

\

\textbf{\textit{Failure type:}} \textcolor{blue}{communication content generation error}

\textbf{\textit{Failure log:}}

\texttt{Bob: low-level-plan:send a message: Hi Alice, I've explored all of the <Kitchen> (5000) and the <Livingroom> (8000), finding <orange> (556740) there. I'm holding <loaf_bread> (12849971). My subplan is to \textcolor{red}{put <loaf_bread> (12849971) into <orange> (556740)}, grasp <orange> (556740), and transport to the bed.}

\textbf{\textit{Analysis:}} \texttt{put <loaf_bread> (12849971) into <orange> (556740)} is inconsistent with common sense and the simulation environment ( Items are randomly initialized in each room) , which is the hallucination of LLM.

\end{tcolorbox}
    \caption{Failure cases of CoBel-World.}
    \label{fig:failure_case}
\end{figure*}
% \section{LLM Usage Disclosure}

% We hereby disclose the use of LLM in the preparation of this manuscript, in compliance with ICLR's submission policies.
% The LLM was  utilized as an assistive tool for language expression refinement during the writing process. Specifically, we leveraged the LLM to optimize the clarity, grammatical accuracy, and writing style of our manuscript. The LLM did not participate in any aspect of research ideation, experimental design or data analysis.
% All content processed with LLM assistance has undergone thorough review, verification, and manual revision by the authors to ensure scientific accuracy, originality, and consistency with the research findings. We confirm that no content generated by the LLM constitutes plagiarism, fabrication of facts, or other forms of scientific misconduct.

% \afterpage{
\begin{figure*}[t]
    \centering
    \begin{tcolorbox}[
    colback=Emerald!10,
    colframe=cyan!40!black,
    title=\textbf{Belief Rules Construction Prompt of Bob}
]
\textbf{Discuss Prompt:} You are bob, you and Alice are constructing belief rules to denote the zero and first order belief of the world. You are required to check the belief rules made by Alice given the challenge description below. Give your reasoning progress in the reasoning:. And then give your comments: Satisfied or Unsatisfied. If Unsatisfied, you should give your suggestions to Alice on how to refine the construction.\\

These suggestions may include:\\
Missing logical relationships among key beliefs, such as omitting the agent's belief about its position.\\
Formatting errors, failing to comply with the prescribed format of the belief language.\\

Belief symbolic language:~\$BELIEF_LANGUAGE\$
Task description:~\$TASK_DESCRIPTION\$
Alice content:~\$ALICE_CONTENT\$
Check if Alice's construction satisfy the need.
Make deletion advice when occurring repeat syntagma.
DO NOT provide suggestions that go beyond the information specified in the task description.\\
Consider ONLY zeroth-order and first-order beliefs.\\
Note that the zeroth-order belief denote my knowledge of the world, first-order belief denote my knowledge of others belief.\\

You should output strictly in the format of the following structure:\\\\
{Reasoning:}\\
{Suggestions:}\\
{Satisfied:(yes or no)}\\

    \end{tcolorbox}
    \caption{Bob's belief rules construction prompt}
    \label{fig:bob_bcp}
\end{figure*}
% }

\begin{figure*}[t]
\centering
\begin{tcolorbox}[
    colback=Emerald!10,
    colframe=cyan!40!black,
    title=\textbf{Belief Rules}
]
zero-order belief rules: \\
\texttt{?agent BELIEVE ?object IN ?room} \\
\texttt{?agent BELIEVE ?bed IN ?room} \\
\texttt{?agent BELIEVE ?container IN ?room} \\
\texttt{?agent BELIEVE ?agent HOLD ?object} \\
\texttt{?agent BELIEVE ?agent HOLD ?container} \\
\texttt{?agent BELIEVE ?container CONTAIN ?object} \\
\texttt{?agent BELIEVE ?room EXPLORED ?exploration\_state} \\
\texttt{?agent BELIEVE ?agent AT ?room} \\[1em]
first-order belief rules: \\
\texttt{?agentA BELIEVE ?agentB BELIEVE ?object IN ?room} \\
\texttt{?agentA BELIEVE ?agentB BELIEVE ?bed IN ?room} \\
\texttt{?agentA BELIEVE ?agentB BELIEVE ?container IN ?room} \\
\texttt{?agentA BELIEVE ?agentB BELIEVE ?agent HOLD ?object} \\
\texttt{?agentA BELIEVE ?agentB BELIEVE ?agent HOLD ?container} \\
\texttt{?agentA BELIEVE ?agentB BELIEVE ?container CONTAIN ?object} \\
\texttt{?agentA BELIEVE ?agentB BELIEVE ?room EXPLORED ?exploration\_state} \\
\texttt{?agentA BELIEVE ?agentB BELIEVE ?agent AT ?room} \\
\end{tcolorbox}
\caption{Illustration of belief rules.}
\label{fig:belief_rules}
% \vspace{-0.2cm}
\end{figure*}

\begin{figure*}[t]
\centering
\begin{tcolorbox}[
    colback=Emerald!10,
    colframe=cyan!40!black,
    title=\textbf{Prompt for First-order Beliefs Update}
]
Assume you are an expert in multi-agent theory-of-mind reasoning. You task is to analyze and extract the information from a multi-agent dialogue. You should take the perspective of \$AGENT_NAME\$ and reason that what information \$OPPO_NAME\$ can get from the dialogue, which is defined as the first-order beliefs and translate these information into structured form. \\

You should follow the steps:
Firstly, extract the information from the dialogue: \\
- What information can \$OPPO_NAME\$ get from the dialogue history(which is defined as the first-order beliefs)? \\
- What \$OPPO_NAME\$ plans to do? \\

Secondly, translate the extracted information (excluding \$OPPO_NAME\$'s plan) into structured first-order beliefs in the form of belief rules without any additional explanation. \\

Notice: \\
1.Maintain the structured beliefs in the format of Belief Rules. \\
2.DO NOT generate information not be mentioned both in Dialogue. \\
3.All entitiess are denoted as <name> (id), such as <table> (712) except the agents' names(e.g. Alice, Bob). \\
4.The exploration state of rooms MUST be part/all/none. \\

Following are provided information for you: \\
Dialogue History: \$MESSAGE\$ \\
Belief Rules: \$RULE\$ \\

Answer strictly in this format: \\
\$OPPO_NAME\$ knows: \\
\$OPPO_NAME\$'s plan: \\
structured first order beliefs:

\end{tcolorbox}
\caption{Prompt for the update of first-order beliefs.}
\label{fig:first-belief-update}
\end{figure*}

\begin{figure*}[t]
\centering
\begin{tcolorbox}[
    colback=Emerald!10,
    colframe=cyan!40!black,
    title=\textbf{Prompt for Zero-Order Beliefs Update}
]

Assume you are a expert good at analyze and extract information from dialogue history. You task is to analyze and extract the information from a multi-agent dialogue and translate these information into structured form. \\
 
You should follow the steps: \\
Firstly, extract the information from the dialogue: \\
- What information can \$AGENT_NAME\$ get from the dialogue history(which is defined as the zero-order beliefs)? \\
- What \$OPPO_NAME\$ plans to do? \\

Secondly, translate the extracted information (excluding \$OPPO_NAME\$'s plan) into structured zero-order beliefs in the form of belief rules without any additional explanation. \\

Notice: \\
1.Maintain the structured beliefs in the format of Belief Rules. \\
2.DO NOT generate information not be mentioned both in Dialogue. \\
3.All entitiess are denoted as <name> (id), such as <table> (712) except the agents' names(e.g. Alice, Bob). \\
4.The exploration state of rooms MUST be part/all/none. \\

Following are provided information for you: \\
Dialogue History: \$MESSAGE\$ \\
Belief Rules: \$RULE\$ \\

Answer strictly in this format: \\
\$AGENT_NAME\$ knows: \\
\$OPPO_NAME\$'s plan: \\
structured zero order beliefs:

\end{tcolorbox}
\caption{Prompt for the zero-order belief update.}
\label{fig:belief-update-zero}
\end{figure*}

\begin{figure*}[t]
\centering
\begin{tcolorbox}[
    colback=Emerald!10,
    colframe=cyan!40!black,
    title=\textbf{Prompt for First-Order Beliefs Prediction}
]
I am \$OPPO_NAME\$. I want to transport as many target objects as possible to the bed with the help of containers. \\

First, please reason over \$OPPO_NAME\$'s state to answer the following question: \\
What the possible locations of goal objects which haven't been transported based on the room exploration state? \\
Goal objects is more likely to be in the rooms which are not fully explored. Put your reasoning behind the 'reasoning:'. Give your analysis in at most two reasons. 

Second, based on your reasoning, please generate the best three plans \$OPPO_NAME\$ will take to transport goal objects as soon as possible. \\

The generated plans must meet following requirements: \\
- One single plan can be broken down into 1 to 3 actions.  \\
- There are 5 allowed actions you can use to construct the plan. \\
1) 'go to': move to a specified room.  
2) 'explore current room <room>(id)': explore current room(is not fully explored) for underlying target objects. 
3) 'go grasp': go to grasp a specified target object.  
4) 'put': Place an object into a specified container. 
5) 'transport': Transport holding objects or containers to the bed and drop them on the bed. \\

Here is an example of a single plan:'go to <Livingroom>(4000), go grasp <apple>(5548447), and transport holding things to the bed', it can be broken down to 3 actions- 'goto <Livingroom>(4000)', 'go grasp <apple>(5548447)' and 'transport holding things to the bed'. \\

Actions take several steps to finish. It may be costly to go to another room or transport to the bed, use these actions sparingly. It will be more efficient to use a container to hold more objects objects and transport to bed at a time. 

Notice: Represent objects, container and room strictly in the format <name>(id) like <livingroom>(1000) <wicker_basket>(5388017). \\

Following are provided information for you: 

Goal: \$GOAL\$ \\
State: \$OPPO_PROGRESS\$ \\
 
What I can do: \\
I can hold two things at a time, and they can be objects or containers. I can grasp ONLY one container at a time and put objects into the holding container to hold more objects at a time. With a container, I can hold at most four objects (three in the container hold by one hand and one object on the other hand). Note that a container can contain three objects, and will be lost once transported to the bed. The room can be explored none/part/all. \\

Answer strictly in this format: \\
reasoning: \\
plans:  \\
plan1:  \\
plan2:  \\
plan3: \\

\end{tcolorbox}
\caption{Prompt for first-order beliefs prediction.}
\label{fig:first-belief-prediction}
\end{figure*}

\begin{figure*}[t]
\centering
\begin{tcolorbox}[
    colback=Emerald!10,
    colframe=cyan!40!black,
    title=\textbf{Prompt for Zero-Order Beliefs Prediction}
]
I am \$AGENT_NAME\$. I want to transport as many target objects as possible to the bed with the help of containers. \\

First, please reason over \$AGENT_NAME\$'s state to answer the following question: \\
Goal objects is more likely to be in the rooms which are not fully explored. Put your reasoning behind the 'reasoning:'. Give your analysis in at most two reasons.

Second, based on your reasoning, please generate one best plan \$AGENT_NAME\$ will take to transport goal objects as soon as possible. \\

The generated plan must meet following requirements: \\
- This plan can be broken down into 1 to 3 actions. \\
- There are 5 allowed actions you can use to construct the plan. \\
1) 'go to': move to a specified room. 
2) 'explore current room <room>(id)': explore current room(is not fully explored) for underlying target objects. 
3) 'go grasp': go to grasp a specified target object. 
4) 'put': Place an object into a specified container. 
5) 'transport': Transport holding objects or containers to the bed and drop them on the bed. \\

Here is an example of a single plan:'go to <Livingroom>(4000), go grasp <apple>(5548447), and transport holding things to the bed', it can be broken down to 3 actions- 'goto <Livingroom>(4000)', 'go grasp <apple>(5548447)' and 'transport holding things to the bed'. \\

Actions take several steps to finish. It may be costly to go to another room or transport to the bed, use these actions sparingly. It will be more efficient to use a container to hold more objects objects and transport to bed at a time. \\

Notice: Represent objects, container and room strictly in the format <name>(id) like <livingroom>(1000) <wicker_basket>(5388017). \\

What I can do: \\
I can hold two things at a time, and they can be objects or containers. I can grasp ONLY one container at a time and put objects into the holding container to hold more objects at a time. With a container, I can hold at most four objects (three in the container hold by one hand and one object on the other hand). Note that a container can contain three objects, and will be lost once transported to the bed. The room can be explored none/part/all. \\

Following are provided information for you:

Goal: \$GOAL\$ 

State: \$MY_PROGRESS\$

Answer strictly in this format: \\
reasoning: \\
plan:
\end{tcolorbox}
\caption{Prompt for zero-order beliefs prediction.}
\label{fig:zero-order-predict}
\end{figure*}

\begin{figure*}[t]
\centering
\begin{tcolorbox}[
    colback=Emerald!10,
    colframe=cyan!40!black,
    title=\textbf{Prompt for Adaptive Collaboration}
]
I am \$AGENT_NAME\$. My teammate \$OPPO_NAME\$ and I want to transport as many target objects as possible to the bed with the help of containers.\\

Please answer the following questions:\\
1. Is there any potential miscoordination between my plan and \$OPPO_NAME\$'s plans or between \$AGENT_NAME\$'s state and \$OPPO_NAME\$'s state? Please analyze the miscoordination in two aspects:

(1) conflicting plans: where my plan and \$OPPO_NAME\$'s plans may conflict in actions or locations.
Such as \$OPPO_NAME\$ and \$AGENT_NAME\$ both plan to explore the same livingroom.

(2) important misaligned information: where some information my state and \$OPPO_NAME\$'s state may misaligned which may lead to miscoordination.

Give your analysis in at most two reasons.\\

2. If there exists heavy miscoordination, please answer Yes; Otherwise, answer No. Allow for a certain degree of information misalignment which can not leads to heavy miscoordination.\\

3. If yes, then please find the misaligned information between my state and \$OPPO_NAME\$'s state. Please list these misaligned pieces of information item by item. Such as I know <apple>(12123). Just list what I know, don't need to list what \$OPPO_NAME\$ knows.\\

4. If no miscoordination, just answer NO.

Following are provided information for you:

\$AGENT_NAME\$'s state: \$MY_PROPGRESS\$

\$OPPO_NAME\$'s state: \$OPPO_PROGRESS\$

\$AGENT_NAME\$'s plan: \$MY_SUBPLAN\$

\$OPPO_NAME\$'s plan: \$OPPO_SUBPLAN\$

Answer in this format:\\
reasons:\\
answer:\\
misaligned information:

\end{tcolorbox}
\caption{Prompt for adaptive collaboration.}
\label{fig:adaptivecollaboration}
\end{figure*}

\begin{figure*}[t]
\centering
\begin{tcolorbox}[
    colback=Emerald!10,
    colframe=cyan!40!black,
    title=\textbf{Prompt for Communication Module}
]
I am \$AGENT_NAME\$. My teammate \$OPPO_NAME\$ and I want to transport as many target objects as possible to the bed with the help of containers. \\

Please help me generate a message to inform \$OPPO_NAME\$ of the misaligned information i know but he don't know and inform \$OPPO_NAME\$ of my plan to achieve our shared goal collaboratively. The message should meet following requirements:

1.The message has to be concise, reliable, and helpful for assisting \$OPPO_NAME\$ and me to collaborate efficiently, and transport as many objects to the bed as possible. \\
2.The message must strictly contain two parts of contents : 1. information only \$AGENT_NAME\$ know and 2. my plan \\

Here is an example of generated massage for you: \\
Example: \\
Message:Hi \$OPPO_NAME\$, I' ve explored all of the <kitchen>(2000) and found <apple>(12123) there. I'm holding <banana>(12234). My plan is to grasp <apple>(12123) and transport holding things to the bed.

Just send what \$AGENT_NAME\$ know, don't need to send what \$OPPO_NAME\$ knows. \\

Following are provided information for you:

Misaligned information: \$MISALIGNED INFORMATION\$

My plan:  \$MY_SUBPLAN\$
\end{tcolorbox}
\caption{Prompt for communication module.}
\label{fig:communication}
\end{figure*}

\begin{figure*}[t]
\centering
\begin{tcolorbox}[
    colback=Emerald!10,
    colframe=cyan!40!black,
    title=\textbf{Prompt for Planning Module}
]
I am \$AGENT_NAME\$. My teammate \$OPPO_NAME\$ and I want to transport as many target objects as possible to the bed with the help of containers. I can hold two things at a time, and they can be objects or containers. I can grasp containers and put objects into them to hold more objects at a time. Actions take several steps to finish. \\

Assume that you are an expert decision maker. Given our shared goal, my plan, my state and previous actions, please analyze the previous action and plan, judge whether the plan has been completed, and if so, respond with 'PLAN DONE'. If plan not be completed, please help me choose the best available action to achieve the goal as soon as possible. Note that a container can contain three objects, and will be lost once transported to the bed. If i'm holding nothing, just grasp a object i found and then keep on exploring or grasping another object.\\

You MUST select a action from the action list.\\

Following are provided information for you: \\
Goal: \$GOAL\$

My plan: \$MY_SUBPLAN\$

Previous action: \$PREVIOUS_ACTIONS\$

My state: \$PROGRESS\$

Action list: \$ACTION_LIST\$ \\

Answer strictly in this format: \\
'answer: your choice'
\end{tcolorbox}
\caption{Prompt for planning module.}
\label{fig:planning}
\end{figure*}

\begin{figure*}[t]
\centering
\begin{tcolorbox}[
    colback=Emerald!10,
    colframe=cyan!40!black,
    title=\textbf{Prompt for Replanning Module}
]
I am \$AGENT_NAME\$. My teammate \$OPPO_NAME\$ and I want to transport as many target objects as possible to the bed with the help of containers. \\

First, please reason over \$AGENT_NAME\$'s state to answer the following question: \\
What the possible locations of goal objects which haven't been transported based on the room exploration state? \\
Goal objects is more likely to be in the rooms which are not fully explored. Put your reasoning behind the 'reasoning:'. Give your analysis in at most two reasons. \\

Second, based on your reasoning and the \$OPPO_NAME\$'s plan, please generate one best plan \$AGENT_NAME\$ will take to transport goal objects as soon as possible while avoiding conflicts with \$OPPO_NAME\$'s plan. The plan should collaborate with \$OPPO_NAME\$ to maximize execution efficiency. \\

The generated plans must meet following requirements: \\
- This plans can be broken down into 1 to 3 actions.  \\
- There are 5 allowed actions you can use to construct the plan. \\
1) 'go to': move to a specified room. 
2) 'explore current room <room>(id)': explore current room(is not fully explored) for underlying target objects. 
3) 'go grasp': go to grasp a specified target object. 
4) 'put': Place an object into a specified container. 
5) 'transport': Transport holding objects or containers to the bed and drop them on the bed. \\

Actions take several steps to finish. It may be costly to go to another room or transport to the bed, use these actions sparingly. \\

It will be more efficient to use a container to hold more objects objects and transport to bed at a time. If i'm holding nothing, just grasp an object i found and then keep on exploring or grasping another object. Avoid transport only one object to bed which cost more time to transport all objects except no more goal objects need to transport. \\

Notice: Represent objects, container and room strictly in the format <name>(id) like <livingroom>(1000) <wicker_basket>(5388017). \\

What I can do:
I can hold two things at a time, and they can be objects or containers. I can grasp ONLY one container at a time and put objects into the holding container to hold more objects at a time. With a container, I can hold at most four objects (three in the container hold by one hand and one object on the other hand). Note that a container can contain three objects, and will be lost once transported to the bed. The room can be explored none/part/all.

Following are provided information for you: \\

Goal: \$GOAL\$

\$OPPO_NAME\$'s plan: \$OPPO_SUBPLAN\$

State: \$MY_PROGRESS\$ \\

Answer strictly in this format: \\
reasoning: \\
plan:
\end{tcolorbox}
\caption{Prompt for replanning module.}
\label{fig:replanning}
\end{figure*}

%%%%%%%%%%%%%%%%%%%%%%%%%%%%%%%%%%%%%%%
\end{document}